\documentclass{article} % For LaTeX2e
\usepackage{iclr2026_conference,times}

\usepackage{amsmath,amsfonts,bm}

\def\eqref#1{equation~\ref{#1}}
\def\1{\bm{1}}

\DeclareMathAlphabet{\mathsfit}{\encodingdefault}{\sfdefault}{m}{sl}
\SetMathAlphabet{\mathsfit}{bold}{\encodingdefault}{\sfdefault}{bx}{n}

\usepackage[breaklinks]{hyperref}
\hypersetup{breaklinks=true, colorlinks=true, urlcolor=blue}
\usepackage[capitalize]{cleveref}
\usepackage{url}
\usepackage{graphicx}
\usepackage{algorithm}
\usepackage{algorithmic}
\usepackage{amsthm}
\usepackage{amssymb}
\usepackage{multirow}
\usepackage{amsmath} 
\usepackage{booktabs}
\usepackage{wrapfig} 
\usepackage{subcaption}
\usepackage{colortbl}
\definecolor{lightblue}{rgb}{0.68,0.90,0.90}

\newtheorem{definition}{Definition}

\newtheorem{proposition}{Proposition}

\title{Brain PathoGraph Learning}
\author{
Ciyuan Peng$^{1}$, Nguyen Linh Dan Le$^{2}$, Shan Jin$^{3}$, Dexuan Ding$^{4}$, Shuo Yu$^{3}$, Feng Xia$^{2}$%\thanks{Corresponding authors: \texttt{f.xia@ieee.org}}\\
\\
$^{1}$Federation University Australia, $^{2}$RMIT University,\\ $^{3}$Dalian University of Technology, $^{4}$Macquarie University \\
}

\iclrfinalcopy % Uncomment for camera-ready version, but NOT for submission.

\begin{document}

\maketitle

\begin{abstract}
%Brain graph learning has demonstrated significant achievements in the fields of neuroscience and artificial intelligence. However, existing methods struggle to selectively learn disease-related knowledge, leading to heavy parameters and computational costs. This challenge diminishes their efficiency, as well as limits their practicality for real-world clinical applications. To this end, we propose a lightweight Brain PathoGraph Learning (BrainPoG) model that enables efficient brain graph learning by focusing solely on pathological patterns and disease-specific characteristics. Specifically, BrainPoG first contains a filter to extract pathological patterns (defined as PathoGraphs) from the whole brain graph, identifying highly disease-relevant abnormal connections. Consequently, less disease-relevant subgraphs are filtered out to achieve graph pruning and lesion localization. Afterwards, a pathological feature distillation module is designed to reduce disease-irrelevant noise features and enhance pathological features of each node. BrainPoG can exclusively learn informative disease-related knowledge while avoiding less relevant information, achieving efficient brain graph learning. Extensive experiments on four benchmark datasets demonstrate that BrainPoG exhibits superiority in both model performance and computational efficiency across various brain disease detection tasks\footnote{The code is available at: \url{https://anonymous.4open.science/r/BrainPoG-6F5E}}.

%Definition of pathological pattern: disease-related abnormal pattern

Brain graph learning has demonstrated significant achievements in the fields of neuroscience and artificial intelligence. However, existing methods struggle to selectively learn disease-related knowledge, leading to heavy parameters and computational costs. This challenge diminishes their efficiency, as well as limits their practicality for real-world clinical applications. To this end, we propose a lightweight Brain PathoGraph Learning (BrainPoG) model that enables efficient brain graph learning by pathological pattern filtering and pathological feature distillation. Specifically, BrainPoG first contains a filter to extract the pathological pattern formulated by highly disease-relevant subgraphs, achieving graph pruning and lesion localization. A PathoGraph is therefore constructed by dropping less disease-relevant subgraphs from the whole brain graph. Afterwards, a pathological feature distillation module is designed to reduce disease-irrelevant noise features and enhance pathological features of each node in the PathoGraph. BrainPoG can exclusively learn informative disease-related knowledge while avoiding less relevant information, achieving efficient brain graph learning. Extensive experiments on four benchmark datasets demonstrate that BrainPoG exhibits superiority in both model performance and computational efficiency across various brain disease detection tasks.%\footnote{The code is available at: \url{https://anonymous.4open.science/r/BrainPoG-6F5E}}.

\end{abstract}

\section{Introduction}

Brain graph learning is pivotal at the intersection of neuroscience and artificial intelligence, aiming to model the complex structural and functional connectivity of the nervous system and to support various clinical tasks~\citep{seguin2023brain,cho2024neurodegenerative,yangwe}. Brain functional graphs~(BFGs), serving as one of the most fundamental representations of brain graphs, provide critical insights into the complex functional connectivity within the brain~\citep{kan2022brain,yu2024long,peng2025biologically}. Typically derived from functional magnetic resonance imaging~(fMRI), BFGs capture functional correlations between the blood-oxygen-level-dependent signals of paired brain regions of interest (ROIs), offering information closely associated with psychiatric and neurological conditions. Therefore, learning from BFGs can support clinical tasks such as the detection of disease-related abnormalities, which is particularly valuable for understanding pathological conditions in the brain~\citep{DBLP:conf/ijcai/Luo000BSMSGY24,DBLP:conf/iclr/XuCDLHBCK25}.

%Brain graph learning has emerged as a critical research topic at the intersection of neuroscience and artificial intelligence, aiming to model the complex structural and functional connectivity of the nervous system and to support various clinical tasks~\citep{seguin2023brain,cho2024neurodegenerative,yangwe}. Brain functional graphs constructed from functional magnetic resonance imaging (fMRI) have attracted increasing attention in many brain graph learning studies for their ability to capture functional interactions between brain regions~\citep{kan2022brain,yu2024long,peng2025biologically}. In such graphs, nodes are regions of interest (ROIs), while edges represent correlations between the blood-oxygen-level-dependent signals of paired ROIs. Owing to its ability to model functional connectivity, fMRI-based brain graph learning has demonstrated great potential in detecting brain dysfunction, which is particularly valuable for disease detection tasks~\citep{DBLP:conf/ijcai/Luo000BSMSGY24,DBLP:conf/iclr/XuCDLHBCK25}.

%1st paragraph: background

Although brain graph learning techniques have achieved initial success, efficiency issue remains a key challenge that limits their practical applicability. Existing methods often indiscriminately learn from the whole brain graph, which contains substantial disease-irrelevant noise, \textit{lacking the ability to pinpoint critical pathological patterns and pathological features}. Particularly, in the absence of precise biomarkers, they often provide insufficient insights into disease-specific alterations in the brain and fail to achieve lesion localization~\citep{mahesh2014biomarker,le2025brainmap}. This limitation not only undermines model performance in downstream tasks but also necessitates larger model architectures to offset the lack of pathological focus. Therefore, learning from complex brain graph data often entails \textit{high parameter demands and substantial computational overhead}~\citep{3599394,zong2024new}. For example, ALTER~\citep{yu2024long} requires 4.6M parameters and 2278MB of memory when trained on the Autism Brain Imaging Data Exchange (ABIDE)\footnote{\url{https://fcon_1000.projects.nitrc.org/indi/abide/}} dataset, while BRAINNETTF~\citep{kan2022brain} uses over 4.0M parameters and 2150MB of memory on ABIDE. Such requirements hinder the applicability of these models and, in particular, limit their practical deployment in time-sensitive or resource-constrained settings.

%2nd: challenge

%To tackle this challenge, we propose Brain PathoGraph Learning (\textbf{BrainPoG}), a lightweight model that explicitly captures pathological patterns and effectively highlights disease-specific characteristics. BrainPoG is designed to avoid learning disease-irrelevant knowledge and instead focus solely on informative pathological patterns and disease-specific characteristics, thereby enabling efficient brain graph learning. 

To tackle this challenge, we propose Brain PathoGraph Learning (\textbf{BrainPoG}), a lightweight model that is designed to avoid learning disease-irrelevant knowledge and instead focus solely on informative pathological patterns and pathological features, thereby enabling efficient brain graph learning. BrainPoG first contains a \textit{pathological pattern filter} to select 
highly disease-relevant subgraphs from the whole brain graph. Specifically, we introduce a subgraph classifier, which evaluates each subgraph with a patho-score, a quantitative measure that evaluates the disease-relevance of a subgraph within the whole brain graph. A high patho-score indicates that the subgraph is highly relevant to the disease, whereas a low patho-score suggests limited relevance. We then introduce \textit{PathoGraph}, which is constructed by dropping less disease-relevant subgraphs (i.e., those with low patho-scores) from the whole brain graph. Therefore, the pathological pattern filter is designed to filter out less disease-relevant subgraphs and form the PathoGraph, achieving graph pruning and lesion localization. Afterwards, we design a \textit{pathological feature distillation module} to drop disease-irrelevant noise features and enhance pathological features of each node in the PathoGraph. This module enables BrainPoG to further focus on disease-specific feature learning, yield more discriminative brain graph representations. Finally, a simple graph convolutional network (GCN)~\citep{kipf2017semi} trained with cross-entropy loss is applied to the PathoGraphs with enhanced node features.

%3rd: proposed work

\textbf{Contributions.}~Building upon the unique design, BrainPoG establishes a lightweight yet effective architecture that not only enables disease-specific feature learning but also substantially reduces model size and computational overhead. This enhances the practicality of BrainPoG for real-world clinical applications. Our main contributions are as follows: \textbf{(1)} We propose a novel Brain PathoGraph Learning (BrainPoG) model, which is designed to enable efficient brain graph learning. \textbf{(2)} We introduce the concept of a PathoGraph and design a pathological pattern filter to achieve graph pruning and facilitating lesion localization. \textbf{(3)} We present a pathological feature distillation method to enhance pathological features, enabling more discriminative brain graph representations. \textbf{(4)} We conduct extensive experiments on four benchmark datasets, demonstrating that BrainPoG outperforms state-of-the-art methods in model performance and computational efficiency.

\section{Related Work}
\paragraph{Brain Graph Learning.}~Brain functional graphs constructed from fMRI reflect the functional connectivity of the human neural system~\citep{bessadok2022graph,qiu2023learning,peng2025biologically,TangMGFHZ24}. In brain functional graphs, nodes represent brain regions of interest (ROIs) and edges are functional correlations between ROIs~\citep{said2024neurograph,cui2022braingb}. With the growing success of graph learning in modeling complex relationships between entities, brain graph learning has emerged as one of the most promising method for various brain graph analytics tasks, such as brain disease detection~\citep{kan2022brain,ding2023lggnet,yu2024long}. A variety of advanced brain graph learning models have been proposed for different objectives~\citep{3599394,gu2025fc,ZhangWYSQGZZ24,TangMGFHZ24}. For example, BrainGNN~\citep{li2021braingnn} captures the functional information of brain graphs constructed from fMRI by ROI-aware graph convolutional layers. STAGIN~\citep{kim2021learning} uses a spatiotemporal attention mechanism in graph neural networks to learn the dynamic brain graph representation, capturing the temporal changes in brain graphs. BRAINNETTF~\citep{kan2022brain} presents a Transformer-based model for brain graph analysis, achieving advanced brain disease detection tasks. MSE-GCN~\citep{lei2023multi} encodes multimodal brain graphs applies by multiple parallel graph convolutional network layers for early Alzheimer’s disease (AD) detection. GroupBNA~\citep{peng2024adaptive} presents a group-adaptive brain network augmentation strategy to construct group-specific brain graphs for accurate brain disease detection. ALTER~\citep{yu2024long} utilizes biased random walks to capture long-range dependencies between brain ROIs. BioBGT~\citep{peng2025biologically} encodes the small-world architecture of brain graphs, therefore enhancing the biological plausibility of the learned brain graph representations.

Although many current brain graph learning methods show promising performance, they often lack the capability to selectively learn disease-related knowledge, compromising efficiency in downstream tasks~\citep{le2025brainmap}. Therefore, they suffer from high parameter requirements and computational overhead, which impedes the practical deployment of these models in time-sensitive or resource-constrained settings. Some works have explored strategies to improve model efficiency. For example, IGS~\citep{3599394} improves model efficiency by iteratively eliminating noisy edges to address the issue of dense brain graphs. However, the challenge of simultaneously maintaining model performance while reducing model size and computational cost remains largely unresolved in brain graph learning. To fill this gap, we propose BrainPoG, a lightweight brain graph learning framework that explicitly learns from pathological patterns and pathological features to achieve efficient brain graph learning.  %making it well-suited for real-world brain disease detection scenarios where efficiency and accuracy must be balanced.

\section{Method}

\begin{figure}
    \centering
    \includegraphics[width=1\linewidth]{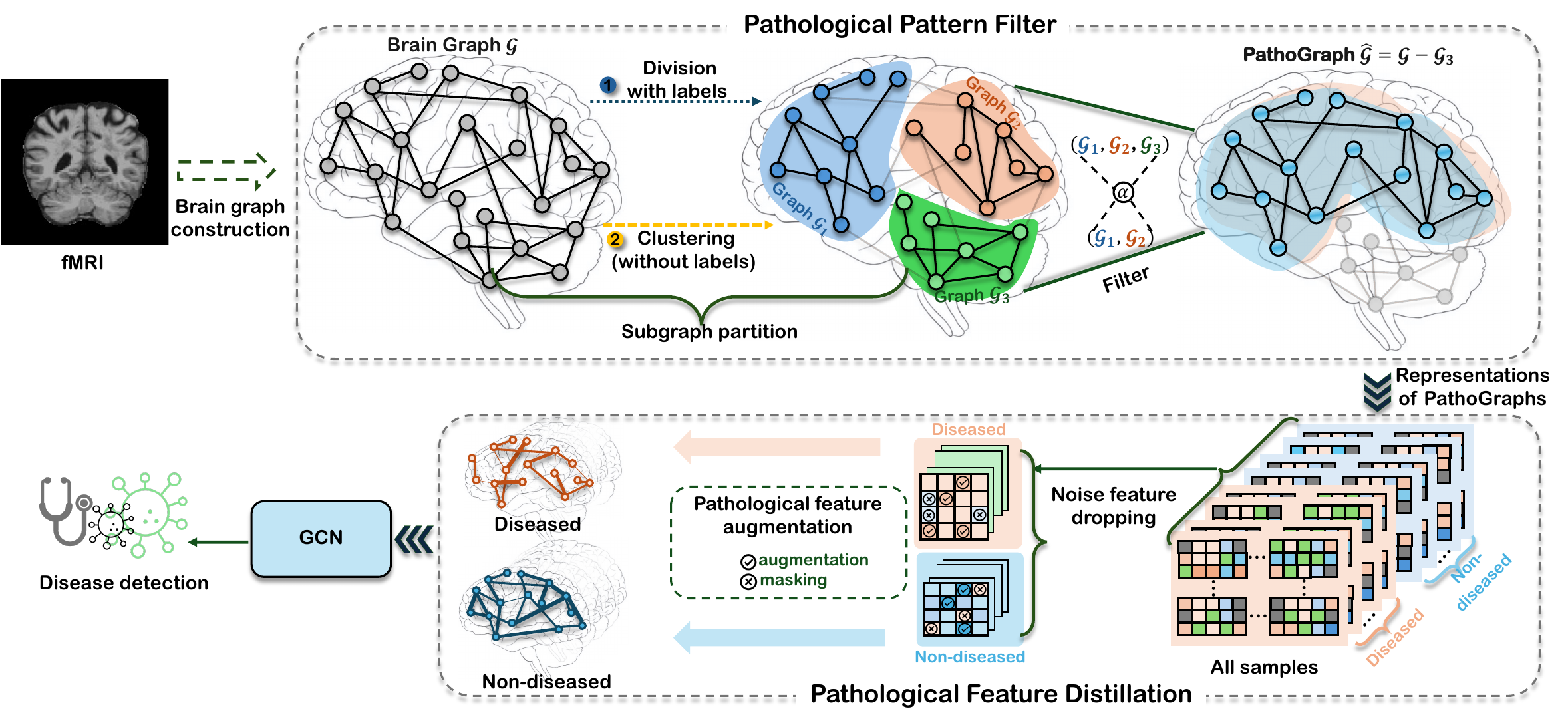}
    \caption{The overall framework of BrainPoG.}
    \label{fig:framework}
\end{figure}

In this section, we introduce our BrainPoG in detail. The overall framework of BrainPoG is illustrated in Figure~\ref{fig:framework}. BrainPoG contains two main modules: (1) pathological pattern filter and (2) pathological feature distillation. The pathological pattern filter is first designed to filter out less disease-relevant subgraphs and construct the PathoGraph. Then, we propose a pathological feature distillation to drop noise features and enhance pathological features of each node in the PathoGraph.

\subsection{Pathological Pattern Filter}

We first construct brain graphs based on fMRI data. Particularly, the functional correlations (e.g., edges between nodes) are the computed Pearson correlation coefficient (PCC)~\citep{cohen2009pearson} between ROIs. For a given brain graph, it is denoted as $\mathcal{G}=(\mathcal{V},\mathbf{A})$, where $\mathcal{V}$ stands for the node (i.e., ROI) set and $\mathbf{A}\in \mathbb{R}^{N\times N}$ is the adjacency matrix. $N$ represents the number of nodes. The node representation $\mathbf{X} \in \mathbb{R}^{N \times N}$ is directly derived from the adjacency matrix, where each node is represented by its corresponding adjacency vector~\citep{kan2022brain}. We first partition the whole brain graph into several subgraphs, each of which represents a functional module of the brain, denoted as $\{\mathcal{G}_1,\mathcal{G}_2,...,\mathcal{G}_m\}$, where $\mathcal{G}_i$ indicates the $i$-th subgraph and $m$ is the number of subgraphs. Particularly, based on different brain parcellation schemes, we adopt different strategies for subgraph partition, as some parcellations provide ROI-level labels while others do not. For graphs with labeled nodes, we partition subgraphs using well-established neuroscience priors. For example, the brain graph constructed upon the AAL atlas~\citep{tzourio2002automated} contains 90 labeled ROIs (i.e., nodes) and can be divided into 9 subgraphs, each representing a distinct functional module. Table~\ref{tab:sub} in Appendix~\ref{appendix_roi_label} shows the detailed subgraph partition for AAL atlas-based brain graphs. In contrast, for graphs without labeled nodes, such as those constructed based on the Craddock 200 atlas~\citep{craddock2012whole}, we obtain subgraphs by applying a community detection method, specifically spectral clustering~\citep{fortunato2010community}, to identify functionally coherent node groups.

Then, we design a filter to extract the pathological pattern formulated by highly disease-relevant subgraphs. Specifically, we implement pathological pattern filtering using a support vector machine (SVM)-based subgraph classifier. We first train an SVM with a radial basis function (RBF) kernel on the brain graph dataset, where each sample corresponds to a complete brain graph $\mathcal{G}$. The SVM outputs a classification score (e.g., the accuracy of graph classification task), denoted as $\alpha$. Afterwards, for each subgraph $\mathcal{G}_i$, we construct a subgraph-only dataset by retaining only $\mathcal{G}_i$ from each sample. This subgraph-specific dataset is then fed into the SVM to obtain the classification score, which is denoted as its patho-score $\beta_i$. Consequently, we can get the set of patho-scores $\beta = \{\beta_1, \beta_2, ..., \beta_m\}$ for all subgraphs, which quantitatively reflect their disease relevance. If $\beta_i \geq \alpha$, subgraph $\mathcal{G}_i$ is considered essential for the disease; otherwise, it is regarded as less relevant. Based on this criterion, we construct the PathoGraph $\hat{\mathcal{G}}$ by dropping subgraphs with lower path-scores (e.g., $\beta_i < \alpha$) and retaining those yielding higher classification performance in the complete brain graph. Here, we give the definition of the PathoGraph.

\begin{definition}[PathoGraph]\label{definition 1}
The PathoGraph $\hat{\mathcal{G}}$ is constructed by dropping less disease-relevant subgraphs from the complete brain graph $\mathcal{G}$, denoted as $\hat{\mathcal{G}}=\mathcal{G}-\{\mathcal{G}\}_{i=1}^{\hat{m}}$. PathoGraph represents the pathological pattern of the brain disease extracted from the complete brain graph.
\end{definition}
Here, $\hat{m}$ indicates the number of dropped subgraphs. By filtering out subgraphs that are less relevant to the disease and obtaining the informative PathoGraph, we not only achieve lesion localization but also effectively achieve graph pruning. Algorithm~\ref{algo1} (in Appendix~\ref{produdure-appendix}) illustrates the procedure of our pathological pattern filter.

\subsection{Pathological Feature Distillation}

\paragraph{Noise Feature Dropping.}
The node representation of PathoGraph $\hat{\mathcal{G}}$ is denoted as $\hat{\mathbf{X}}\in \mathbb{R}^{\hat{N}\times\hat{N}}$, where $\hat{N}  \ll N$ is the number of retained nodes after filtering. For all $D$ samples in the PathoGraph dataset $\hat{\mathcal{B}}$, we use singular value decomposition (SVD)-based scoring function to identify communal features across all samples. Because communal features generally exist across both healthy and diseased subjects, they are disease-irrelevant features. Therefore, communal features are regarded as noise features and subsequently dropped to reduce redundancy.

Given the node representation set of $\hat{\mathcal{B}}$,
$\{\hat{\mathbf{X}}^{(1)},\hat{\mathbf{X}}^{(2)},...,\hat{\mathbf{X}}^{(D)}\}$, we first rearrange the features of each node across all $D$ subjects to construct cross-subject feature representations. Formally, for node $v_i$, its cross-subject feature representation is $\mathbf{C}_{v_i}\in \mathbb{R}^{\hat{N}\times D}$. Then, we perform SVD on the cross-subject feature representations to evaluate the significance of each feature dimension of each node. The first column of the left singular matrix obtained from SVD is used as the scoring criterion for identifying communal features across all subjects. Therefore, a communal feature score matrix $\mathbf{S}\in \mathbb{R}^{\hat{N}\times\hat{N}}$ for all nodes can be obtained:
\begin{equation}\label{eq1}
\begin{split}
\mathbf{C}_{v_i} &= \mathbf{L}_{v_i} \mathbf{\Sigma}_{v_i} \mathbf{R}_{v_i}^{\top}, \quad v_i \in \hat{N}, \\
\mathbf{l}_{v_i} &= \mathbf{L}_{v_i}[:,1] = 
\begin{bmatrix}
\mu^1_{v_i}, \mu^2_{v_i}, \ldots, \mu^{\hat{N}}_{v_i}
\end{bmatrix}^{\top}, \\
\mathbf{S} &= [\mathbf{l}_{v_1}, \mathbf{l}_{v_2}, \ldots, \mathbf{l}_{v_{\hat{N}}}]^{\top}.
\end{split}
\end{equation}
Here, $\mathbf{L}_{v_i}$ and $\mathbf{R}_{v_i}$ indicate the left and right singular matrices of node $v_i$, respectively, and $\mathbf{\Sigma}_{v_i}$ is the diagonal matrix. $\mathbf{l}_{v_i}$ is the first left singular vector and stands for the feature score vector of node $v_i$, where $\mu^{j}_{v_i}$ is the score of the $j$-th feature. To identify communal features across all subjects, we rank the features of each node based on their feature scores. 

\begin{proposition}[Communal features]\label{proposition1}
Based on feature scores, the top ranked features in the cross-subject feature representation $\mathbf{C}_{v_i}$ are considered as communal features of node $v_i$ across all subjects.
\end{proposition}

The proof of Proposition~\ref{proposition1} is given in Appendix~\ref{proof1}. Specifically, the top-$k$ ranked features are considered as communal features across all subjects and are dropped to achieve feature distillation and eliminate redundancy. In addition, we also drop the bottom-$k$ features due to their limited contributions to informative node representations. Afterwards, for the PathoGraph $\hat{\mathcal{G}}$, we can get its distilled node representation, denoted as $\widetilde{\mathbf{X}}\in \mathbb{R}^{\hat{N}\times \hat{N}^{\prime}}$, where $\hat{N}^{\prime}  \ll \hat{N}$ is determined by the value of $k$.

\paragraph{Pathological Feature Augmentation.}~After dropping noise features, we further enhance the pathological features (e.g., group-specific features within each subject group) to strengthen their discriminative representations. For example, in the Attention Deficit Hyperactivity Disorder (ADHD-200) dataset, samples can be divided into the normal control group and the ADHD patient group. We separately enhance pathological features in the ADHD patient group and features in the NC group, highlighting the disease-specific features.

For the dataset $\hat{\mathcal{B}}$, subjects are divided into $Y$ groups based on the given labels. We apply the same SVD-based scoring function with Equation~(\ref{eq1}) to the node representations of each group independently, obtaining the group-specific feature score matrices $\{\mathbf{F}_1,\mathbf{F}_2,...,\mathbf{F}_Y\}$, where $\mathbf{F}_y \in \mathbb{R}^{\hat{N}\times \hat{N}^{\prime}}$ is the group-specific feature score matrix for group $y$. Features with high group-specific scores are considered as important pathological features. Then, we introduce a feature augmentation strategy based on group-specific feature score matrices. Compared to existing feature augmentation methods~\citep{zhu2021graph,zhang2023spectral}, our approach incorporates a feature masking operation guided by the group-specific feature scores. Specifically, features with higher group-specific scores are retained, while less informative features are masked out to enhance group-discriminative representation. The feature weights are defined as:
\begin{eqnarray}\label{eq2}
\mathbf{W}=\frac{1}{\hat{N}}\sum_{i=1}^{\hat{N}}|\widetilde{\mathbf{x}}_i|\odot \mathbf{F}_{y,i},
\end{eqnarray}
where $\widetilde{\mathbf{x}}_i\in \mathbb{R}^{\hat{N}^{\prime}}$ is the feature vector of node $v_i$, and $\mathbf{F}_{y,i}\in \mathbb{R}^{\hat{N}^{\prime}}$ represents the group-specific score vector of node $v_i$ in group $y$. $\odot$ denotes element-wise multiplication. $\mathbf{W}=[w_1,w_2,...,w_{\hat{N}^{\prime}}]$ contains the weights for each feature dimension. Then, for the $j$-th feature, we calculate the masking probability according to its importance (weight) obtained from Equation~(\ref{eq2}):
\begin{eqnarray}
p_j=\min\left(\rho\cdot\frac{\log w_{max}-\log w_j}{\log w_{max}-\log \lambda_w},p_t\right).
\end{eqnarray}  
Here, $\rho$ is a hyperparameter adjusting the magnitude of feature enhancement, $w_{max}$ and $\lambda_w$ are the maximum and average weight values of all feature dimensions, and $p_t<1$ is a truncation threshold that prevents overly aggressive deletion. We then sample a binary mask vector $\tilde{\mathbf{b}}\in\{0,1\}^{\hat{N}^{\prime}}$, where each dimension element is sampled from a Bernoulli distribution:
\begin{equation}
\tilde{b}_j \sim \text{Bernoulli}(1 - p_j), \quad \forall j = 1, 2, \ldots, \hat{N}'.
\end{equation}
By sampling the binary mask from the Bernoulli distribution, we can more robustly identify informative features~\citep{lee2022self}.
Finally, the enhanced node representation is computed as:  
\begin{eqnarray}    \widetilde{\mathbf{X}}^{\prime}=\left[\widetilde{\mathbf{x}}_1\odot\tilde{\mathbf{b}};\widetilde{\mathbf{x}}_2\odot\tilde{\mathbf{b}};...;\widetilde{\mathbf{x}}_N\odot\tilde{\mathbf{b}}\right]^{\top}.
\end{eqnarray}
Algorithm~\ref{algo2} (in Appendix~\ref{produdure-appendix}) shows the procedure of the pathological feature distillation. After obtaining the enhanced node representations of all PathoGraphs, we fed them into a simple GCN model. Specifically, we employ the cross-entropy loss function to optimize the GCN model.

\section{Experiments}

\subsection{Experimental Setup}

\paragraph{Datasets.}
We conduct experiments on fMRI data collected from four benchmark datasets. (1) Alzheimer’s Disease Neuroimaging Initiative (ADNI)~\footnote{\url{https://adni.loni.usc.edu/}} dataset contains 407 subjects, including 190 normal controls (NCs), 170 mild cognitive impairment (MCI) patients, and 47 Alzheimer’s disease (AD) patients. (2) Parkinson's Progression Markers Initiative (PPMI) dataset~\citep{marek2018parkinson} includes fMRI data of $49$ Parkinson's disease (PD) patients, $69$ individuals at risk for PD (Prodromal), and $40$ NCs. (3) Attention Deficit Hyperactivity Disorder (ADHD-200)~\footnote{\url{https://fcon_1000.projects.nitrc.org/indi/adhd200/}} dataset comprises a total of 459 subjects in which 230 subjects are typically developing individuals and 229 subjects are ADHD patients. (4) Autism Brain Imaging Data Exchange (ABIDE)~\footnote{\url{https://fcon_1000.projects.nitrc.org/indi/abide/}} dataset contains 1, 009 subjects including 516 Autism spectrum disorder patients and 493 NCs. The definition of ROI in ADNI and PPMI datasets is based on AAL atlas~\citep{tzourio2002automated}. Therefore, graphs in these two datasets have ROI-level labels, and their subgraphs can be obtained by division with labels (see Table~\ref{tab:sub}). The ROIs of graphs in ADHD-200 and ABIDE datasets are defined by Craddock 200 atlas~\citep{craddock2012whole}, which does not provide ROI-level labels. Thereby, we apply spectral
clustering to generate subgraphs for these two datasets. The number of ROIs in ADNI and PPMI is 90. For ABIDE and ADHD-200, the numbers of ROIs are 200 and 190, respectively.

\paragraph{Evaluation Metrics.}

We evaluate our model on the disease detection task, which is regarded as graph classification problem. Notably, for ABIDE and ADHD-200 datasets, the classification task is a binary classification problem, detecting whether the subject is a patient or a normal control. For ADNI and PPMI datasets, which have three groups, the disease detection is a multiple classification problem. To evaluate the model performance, we choose three metrics, including accuracy (ACC), F1 score, area under the receiver operating characteristic curve (AUC). For the multiclass classification task on ADNI and PPMI datasets, we use macro averaging for the F1 score. In addition, to evaluate the model efficiency, we calculate the parameter number, running time, and memory usage. %All results are the average values of 10 random runs on test sets with the standard deviation.

\paragraph{Baseline Methods.}

We compare our model with state-of-the-art brain graph learning models, including BrainGB~\citep{cui2022braingb}, BRAINNETTF~\citep{kan2022brain}, A-GCL~\citep{zhang2023gcl}, MCST-GCN~\citep{zhu2024spatio}, ALTER~\citep{yu2024long}, BioBGT~\citep{peng2025biologically}, BrainOOD~\citep{DBLP:conf/iclr/XuCDLHBCK25}. In addition, we compare our method with two representative graph learning baselines, GCN~\citep{kipf2017semi} and graph attention network (GAT)~\citep{velickovic2018graph}, as well as two typical machine learning approaches, SVM and random forest (RF).

\paragraph{Implementation Details.}

Our model is implemented using PyTorch Geometric v2.6.1, PyTorch v2.7.0, and NetworkX v3.4.2. Model training is performed on an NVIDIA A10G GPU with 64GB of memory. We train our model with the Adam optimizer and use cross-entropy loss for classification. For evaluation, we perform stratified five-fold cross-validation, in each fold, the dataset is randomly partitioned into 70\% training, 10\% validation, and 20\% test sets. Full implementation is given in Appendix~\ref{imple}.

\subsection{Results}

\paragraph{Model Performance Comparison.}
Table~\ref{Tab:comparesion_result} compares the results of BrainPoG with baselines. As shown by the experimental results, BrainPoG consistently outperforms other baselines across all four datasets. Notably, BrainPoG achieves the highest accuracy of 83.31\%, 82.90\%, 93.16\%, and 93.03\% on four datasets, respectively, representing improvements of 7.88\%, 2.03\%, 8.83\%, and 6.91\% over the second-best methods. Given the binary-classification nature of ABIDE and ADHD-200, higher accuracy is expected on those two datasets. Beyond accuracy, BrainPoG also demonstrates the highest AUC and F1 across four datasets, achieving AUCs of 92.23\%, 92.00\%, 98.92\% and 97.77\%, and F1 scores of 69.86\%, 81.22\%, 93.13\%, and 93.03\%, respectively, indicating large gains in class-wise discrimination. The experimental results demonstrate that our model excels in various brain disease detection tasks.

\begin{table*}[ht]
	\centering
	\caption{Experimental results on four datasets (\%). The best results are marked in \textbf{bold}, and the suboptimal results are marked \underline{underlined}.}
	\label{Tab:comparesion_result}
	\renewcommand\arraystretch{1}
    \setlength{\tabcolsep}{3pt} % Adjust column separation	
    \small
    \resizebox{\textwidth}{!}{
    \begin{tabular}{cccc|ccc|ccc|ccc}
    \toprule
    \multirow{2}{*}{Method} & \multicolumn{3}{c|}{ADNI} &\multicolumn{3}{c|}{PPMI} &\multicolumn{3}{c|}{ABIDE} &\multicolumn{3}{c}{ADHD-200} \\\cmidrule{2-13}
    & ACC & AUC &F1 &ACC & AUC&F1 & ACC & AUC &F1& ACC & AUC&F1\\
    \midrule
    SVM& 51.37$\pm$2.81 & 61.02$\pm$3.21&35.28$\pm$2.03&41.00$\pm$4.06&55.04$\pm$3.16&19.47$\pm$1.30&57.00$\pm$7.65&71.73$\pm$2.58&50.43$\pm$11.00 &58.20$\pm$7.66&70.10$\pm$2.55&51.21$\pm$7.33 \\
    
    RF& 55.69$\pm$4.45 & 62.76$\pm$5.54&39.28$\pm$3.13
&37.50$\pm$6.89&49.42$\pm$5.80&27.84$\pm$8.72
&53.04$\pm$3.74&56.36$\pm$3.97&52.93$\pm$3.80
&65.22$\pm$2.64&75.49$\pm$1.64 &65.16$\pm$2.65\\
    \midrule
    GCN& 51.43$\pm$1.90&55.31$\pm$3.39&35.49$\pm$1.35
    &45.88$\pm$11.41&58.24$\pm$6.14&32.76$\pm$4.66
    &55.29$\pm$4.87&68.23$\pm$2.57&53.82$\pm$5.00
    &62.13$\pm$7.17&74.09$\pm$5.73&61.10$\pm$7.37
    
    \\
    GAT& 51.95$\pm$5.43&61.18$\pm$4.44&39.38$\pm$6.47
    &55.00$\pm$1.53&63.72$\pm$3.46&44.99$\pm$5.79
    &58.71$\pm$1.99&60.98$\pm$3.40&57.05$\pm$2.56
    &59.35$\pm$4.22&63.37$\pm$4.69&58.66$\pm$4.01
    
    \\
    \midrule
    BrainGB& 54.88$\pm$2.44&65.16$\pm$3.66&43.92$\pm$3.47
    &31.87$\pm$11.76&50.00$\pm$4.10&37.81$\pm$11.66
    &62.87$\pm$1.00&70.42$\pm$8.40&62.94$\pm$2.20
    &65.22$\pm$1.23&71.34$\pm$5.30&65.15$\pm$7.20    
    \\
    BRAINNETTF & 70.43$\pm$3.21&77.65$\pm$2.42&57.54$\pm$2.87
    &\underline{80.87$\pm$4.01}&\underline{87.43$\pm$3.01}&77.30$\pm$3.45
    &77.33$\pm$1.04&84.27$\pm$2.18&76.86$\pm$4.23
    &75.12$\pm$1.67&82.56$\pm$2.34&75.03$\pm$1.09
    \\
     A-GCL& 56.43$\pm$2.12&60.12$\pm$2.23&47.76$\pm$3.56
     &57.54$\pm$2.87&61.23$\pm$2.45&56.12$\pm$3.76
     &62.65$\pm$2.88&68.42$\pm$2.85&62.14$\pm$2.96
     &65.12$\pm$4.45&70.22$\pm$4.28&64.79$\pm$4.51
     
     \\     
     MCST-GCN & 
     \underline{75.43$\pm$2.54} &\underline{80.65$\pm$2.10}&61.54$\pm$2.81
     &79.87$\pm$4.01& 85.43$\pm$2.94&\underline{77.34$\pm$3.40}
     &\underline{84.33$\pm$1.04}&\underline{87.10$\pm$2.18}&\underline{84.26$\pm$4.23}
     &\underline{86.12$\pm$1.67}&\underline{91.56$\pm$2.34} & \underline{85.73$\pm$1.09} \\

     ALTER & 66.38$\pm$5.51 & 77.65$\pm$4.58 & \underline{68.87$\pm$4.71} & 64.29$\pm$5.05 & 83.38$\pm$2.66 & 71.36$\pm$8.62 & 71.20$\pm$1.30 & 76.60$\pm$1.13 & 72.43$\pm$1.55 & 74.05$\pm$8.02 & 82.62$\pm$5.63 & 74.55$\pm$6.84 \\ 
     
     BioBGT  &55.11$\pm$2.34 & 62.48$\pm$5.02 &53.89$\pm$2.64
     &54.57$\pm$4.69& 65.20$\pm$6.03 & 67.20$\pm$1.76
     &74.41$\pm$2.32&73.33$\pm$2.34 &68.43$\pm$2.11
     &71.90$\pm$1.28&70.75$\pm$1.76  & 75.31$\pm$1.06 \\ 
     
     BrainOOD  &68.10$\pm$3.97 & 69.84$\pm$3.89 &65.23$\pm$5.35
     &66.10$\pm$5.42& 64.93$\pm$6.02 &62.12$\pm$3.33
     &66.24$\pm$5.12&64.21$\pm$3.29&63.89$\pm$4.20
     &67.43$\pm$4.47&66.59$\pm$3.75&65.39$\pm$4.11 \\ 
     \midrule
     \textbf{BrainPoG} & \textbf{83.31$\pm$4.90}& \textbf{92.23$\pm$2.83}&\textbf{69.86$\pm$8.21}
     &\textbf{82.90$\pm$1.62}&\textbf{92.00$\pm$1.50}&\textbf{81.22$\pm$1.66}
     &\textbf{93.16$\pm$2.46}&\textbf{98.92$\pm$0.67}&\textbf{93.13$\pm$2.49}
     &\textbf{93.03$\pm$3.27}&\textbf{97.77$\pm$2.19}&\textbf{93.03$\pm$3.27}\\
    \bottomrule
    \end{tabular}}
\end{table*}

\paragraph{Model Efficiency Analysis.}
In addition to evaluating model performance, it is crucial to assess the computational efficiency of the proposed method. To this end, we report the number of parameters, running time, and memory usage of BrainPoG during training across all selected datasets. As summarized in Table~\ref{Tab:efficiency_result}, BrainPoG requires fewer parameters, shorter running time, and lower computational overhead than state-of-the-art brain graph learning models. By contrast, simple models such as SVM, RF, and GCN indeed use a few of parameters and computational costs but suffer from significantly worse model performance, whereas methods that achieve accuracy comparable to BrainPoG demand more parameters and computational overhead. These results demonstrate that BrainPoG yields substantial improvements in computational efficiency while improving model performance. In addition, to further verify the efficiency of BrainPoG, we compare its time and space complexity with other state-of-the-art brain graph learning models (see Table~\ref{tab:complexity} in Appendix~\ref{appendix_efficiency}).

\begin{table*}[ht]
	\centering
	\caption{ Model efficiency comparison.}
	\label{Tab:efficiency_result}
	\renewcommand\arraystretch{1}
    \setlength{\tabcolsep}{3pt} % Adjust column separation	
    \small
    \resizebox{\textwidth}{!}{
    \begin{tabular}{cccc|ccc|ccc|ccc}
    \toprule
    \multirow{2}{*}{Method} & \multicolumn{3}{c|}{ADNI} &\multicolumn{3}{c|}{PPMI} &\multicolumn{3}{c|}{ABIDE} &\multicolumn{3}{c}{ADHD-200} \\\cmidrule{2-13}
    & Parameter No. & Running time&Memory&Parameter No. & Running time&Memory& Parameter No. & Running time&Memory& Parameter No. & Running time&Memory\\
    \midrule
    SVM&579&0.0332s/run&0.37MB&239&0.0079s/run&0.15MB &238&0.0078s/run&0.14MB &238&0.0078s/run&0.14MB\\
    
    RF&11K&0.4463s/run&0.24MB & 7K&0.3026s/run&0.13MB & 20K&1.0445s/run&0.84MB & 9K&0.5158s/run & 0.40MB\\
    \midrule
    GCN& 8K&8.6810s/epoch&74MB & 8K&5.9670s/epoch&26MB & 15K&36.1985s/epoch&698MB & 14K&16.8647s/epoch&424MB    
    \\
    GAT& 52K&0.8165s/epoch&1123MB & 52K&0.4017s/epoch&1182MB & 108K&1.9470s/epoch & 1278MB & 103K&0.8945s/epoch&1186MB
    
    \\
    \midrule
    BrainGB& 628K&2.7192s/epoch&1219MB&628K&1.1525s/epoch&1229MB&981K&31.2101s/epoch&1807MB&898K&11.6271s/epoch&1517MB
    \\
    BRAINNETTF & 1.8M&10.2300s/epoch&1752MB&1.8M&6.3400s/epoch&1700MB&4.0M&42.5600s/epoch&2150MB&3.8M&18.7623s/epoch&1970MB
    \\
     A-GCL& 246K&3.5060s/epoch&450MB&246K&1.5543s/epoch&380MB&468K&8.5600s/epoch&750MB&448K&5.2800s/epoch&620MB
     
     \\     
     MCST-GCN &1.3M&5.8392s/epoch&1550MB & 1.3M&5.0323s/epoch&1522MB &3.8M&39.8904s/epoch&2088MB & 3.7M & 15.9895s/epoch&1788MB 
      \\

     ALTER &  1.6M & 9.2993s/epoch & 1580MB & 1.4M & 4.8729s/epoch & 1510MB & 4.6M & 44.5332s/epoch & 2278MB & 4.3M & 23.0471s/epoch & 2109MB \\ 
     
     BioBGT  &455K&8.5322s/epoch&1023MB&338K&3.2800s/epoch&422MB&465K&15.300s/epoch&917MB& 454K&5.2150s/epoch&590MB
     \\ 
     
     BrainOOD  &469K&6.3000s/epoch&846MB&391K&2.5810s/epoch&380MB&528K&11.4728s/epoch&872MB&496K&6.4000s/epoch&574MB
      \\ 
     \midrule
     \rowcolor{lightblue}\textbf{BrainPoG} & 227K& 0.0046s/epoch&23MB
     &11K&0.0029s/epoch&17MB
     &415K&0.0030s/epoch&99MB
     &140K&0.0030s/epoch&23MB
     \\
    \bottomrule
    \end{tabular}}
\end{table*}

\paragraph{PathoGraph Identification and Lesion Localization.}

\begin{figure*}[ht]
    \centering
    \begin{subfigure}{0.24\textwidth}
        \includegraphics[width=\linewidth]{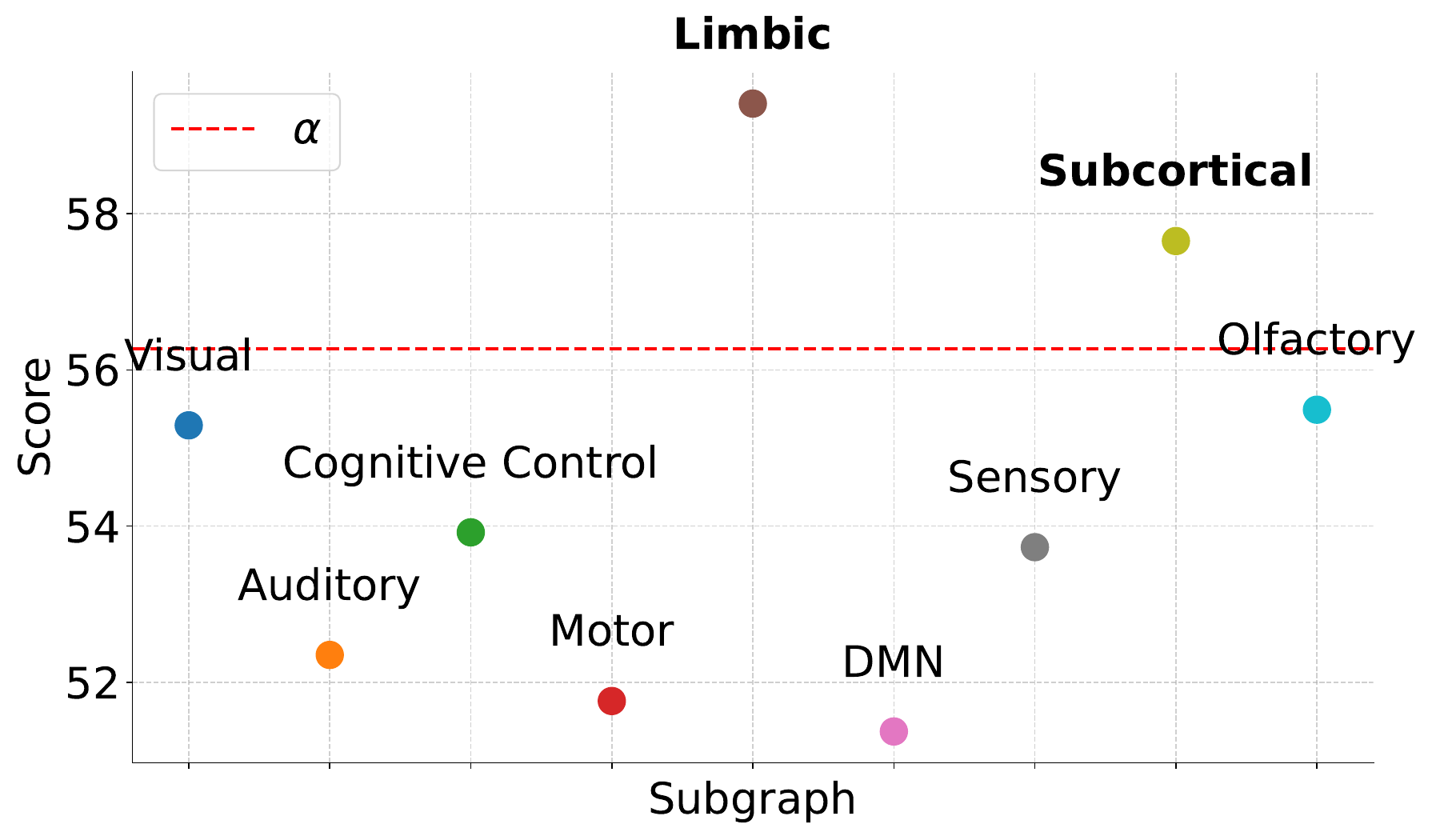}
        \caption{ADNI}
        \label{fig:sub1_patho}
    \end{subfigure}
   % \hfill
    \begin{subfigure}{0.24\textwidth}
   % \vspace{-1\baselineskip}
        \includegraphics[width=\linewidth]{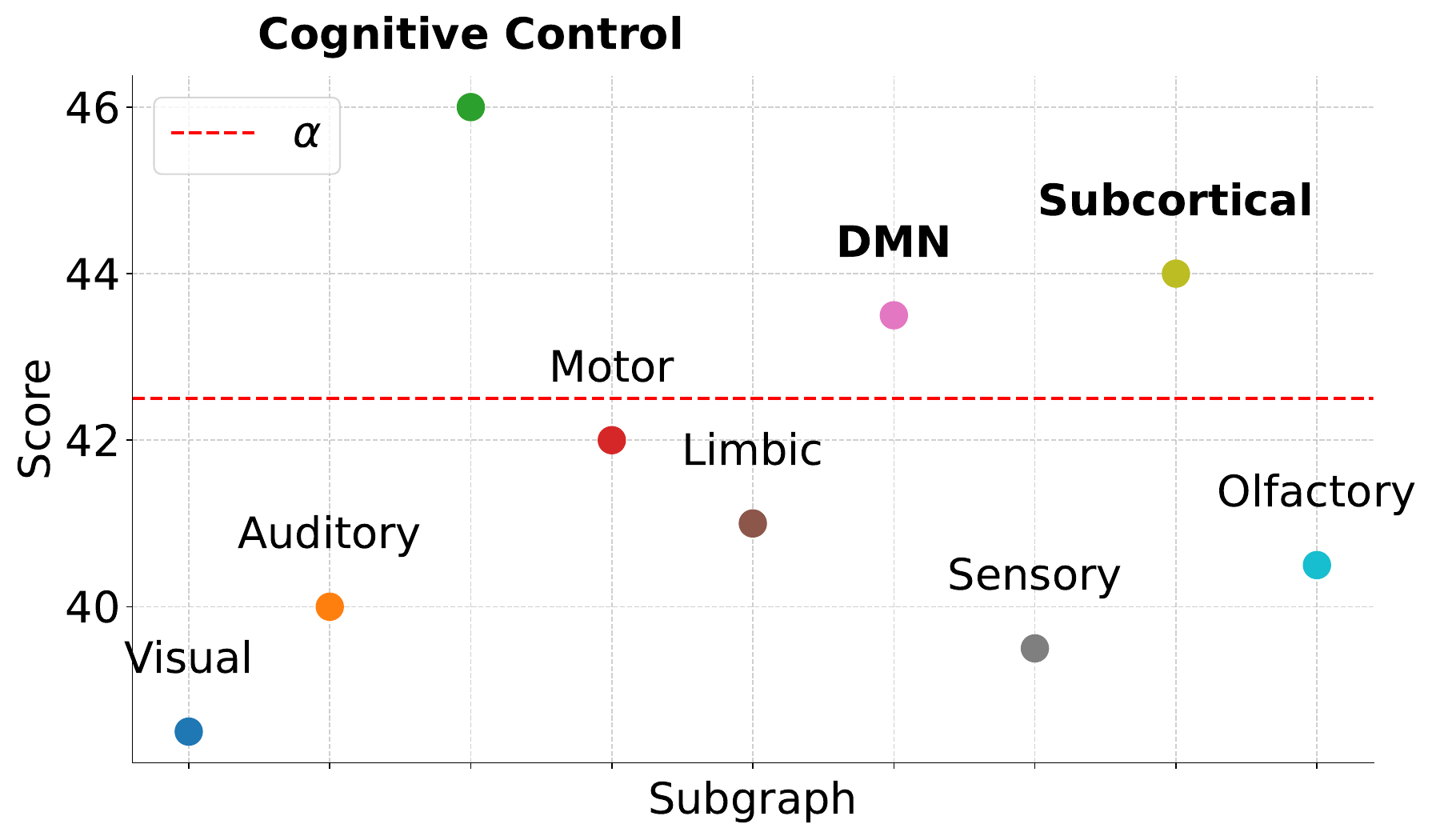}
        \caption{PPMI}
        \label{fig:sub2_patho}
    \end{subfigure}
   % \hfill
    \begin{subfigure}{0.24\textwidth}
        \includegraphics[width=\linewidth]{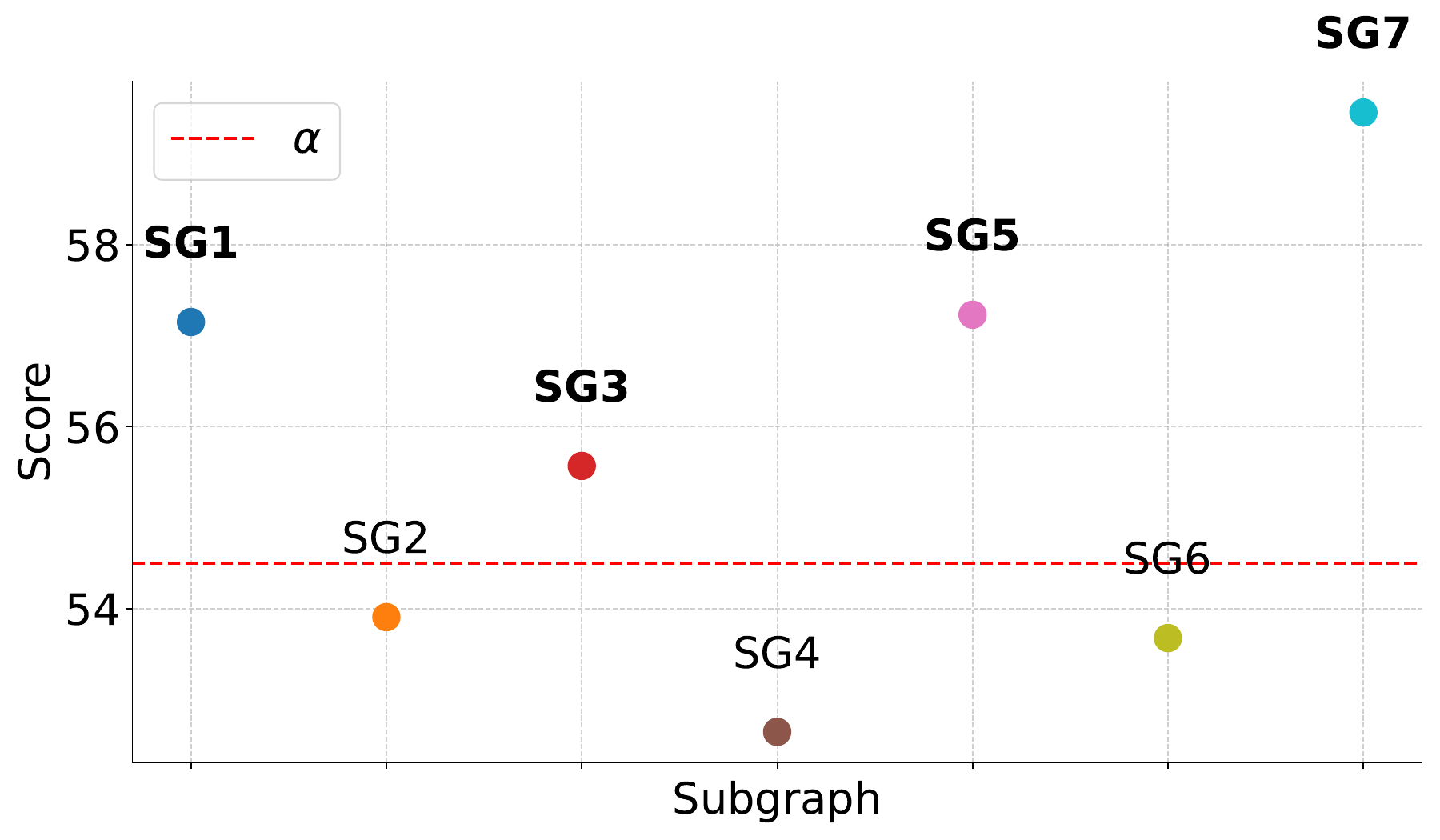}
        \caption{ABIDE}
        \label{fig:sub3_patho}
    \end{subfigure}
    \begin{subfigure}{0.24\textwidth}
        \includegraphics[width=\linewidth]{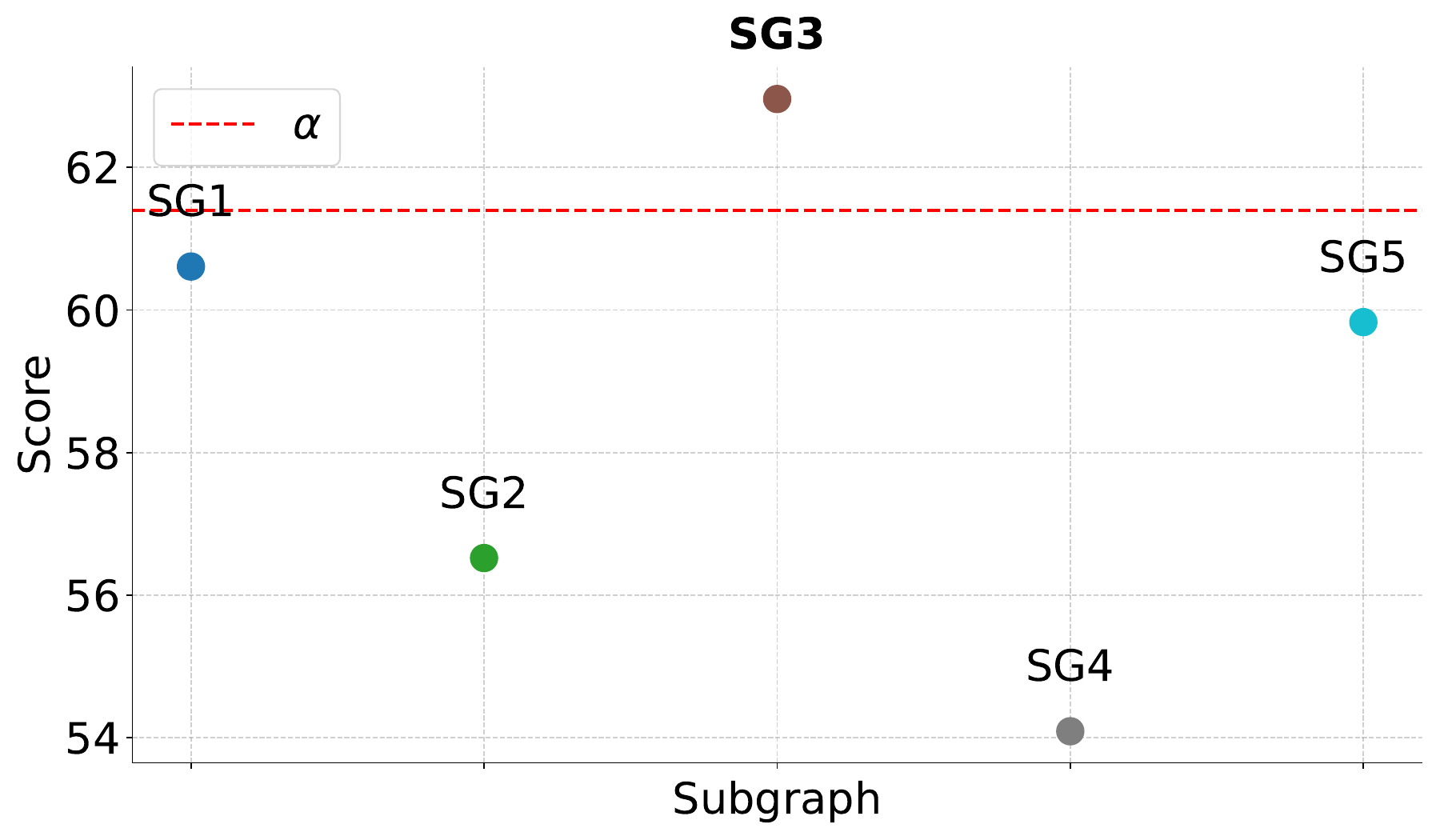}
        \caption{ADHD-200}
        \label{fig:sub4_patho}
    \end{subfigure}
    \caption{PathoGraph identification for four different diseases. The PathoGraph is constructed by remaining subgraphs shown in \textbf{bold} and dropping those having lower patho-scores than $\alpha$ . `SG' denotes the subgraph without labels.}
   % \vspace{-1\baselineskip} 
    \label{fig:AD_patho_identify}
\end{figure*}

\begin{figure*}[ht]
    \centering
    \begin{subfigure}{0.4\textwidth}
        \includegraphics[width=1\linewidth]{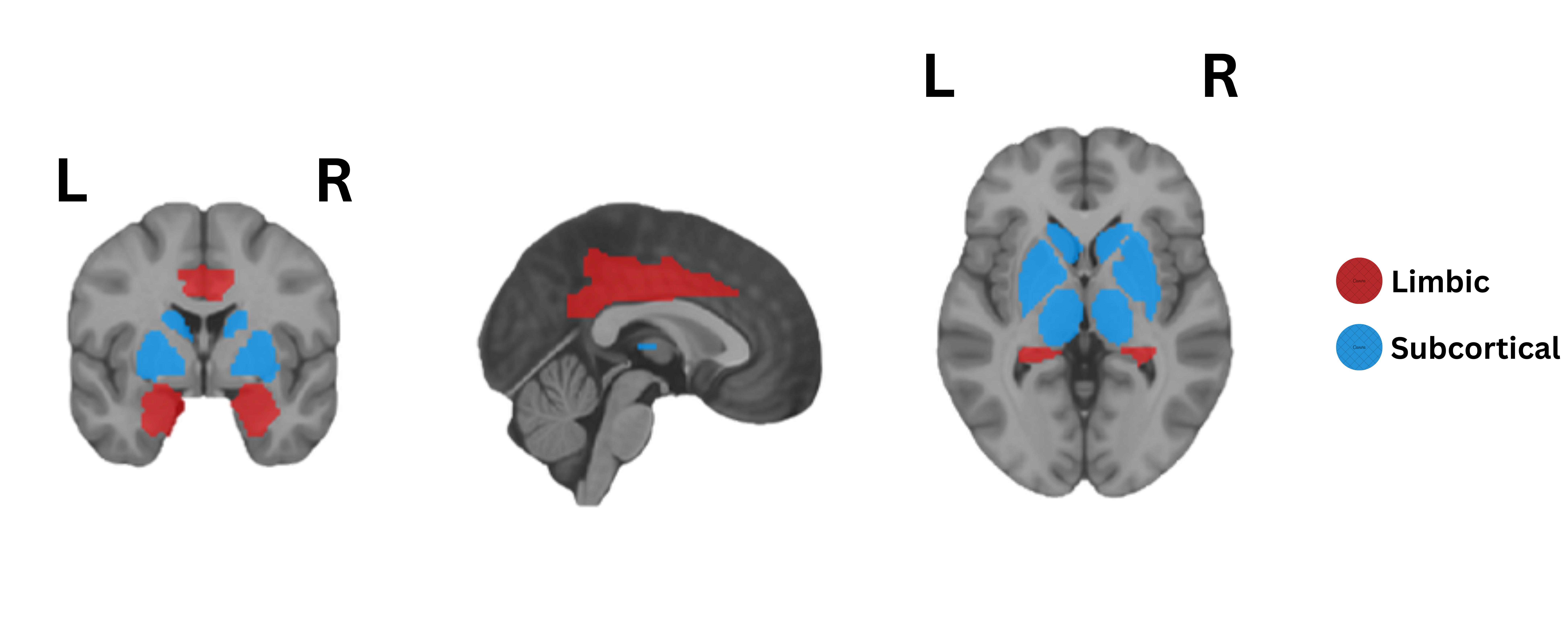}
        \caption{ADNI}
        \label{fig:sub1_lesion}
    \end{subfigure}
   % \hfill
    \begin{subfigure}{0.4\textwidth}
        \includegraphics[width=1\linewidth]{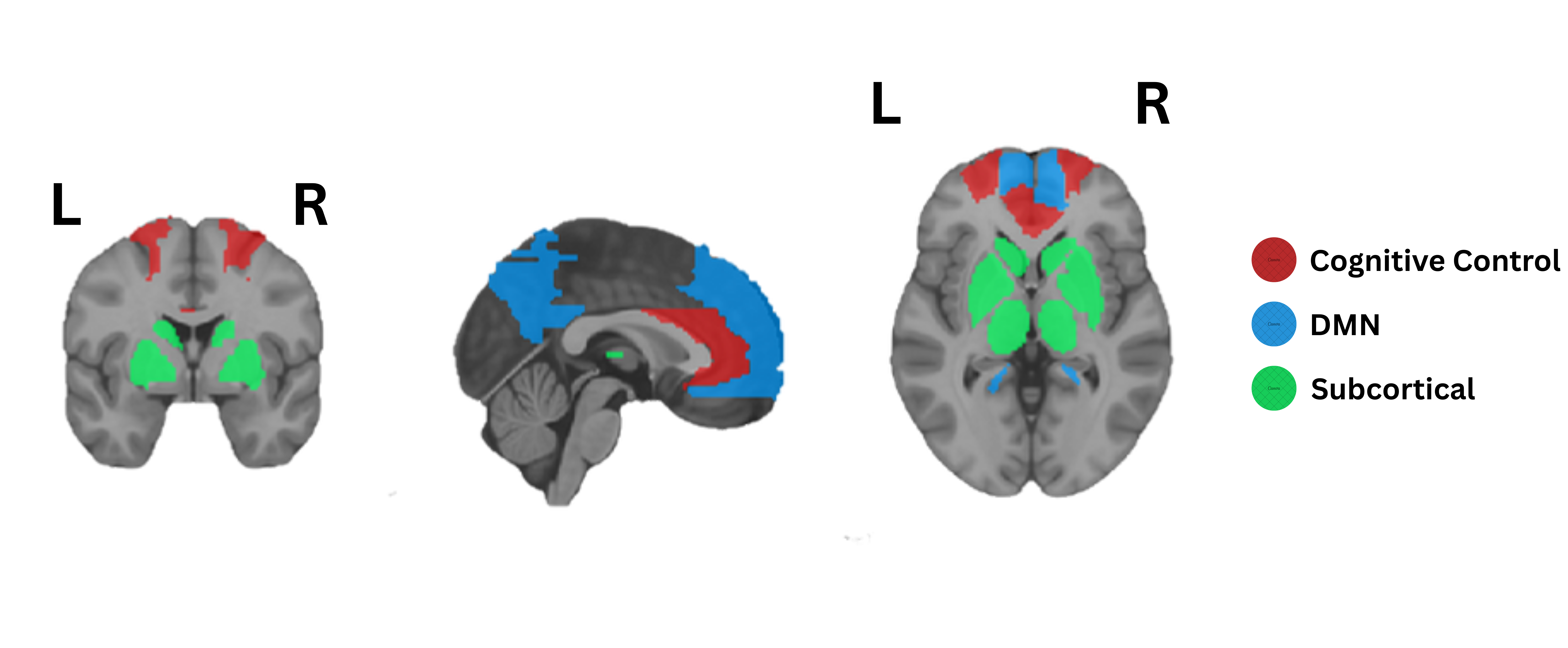}
        \caption{PPMI}
        \label{fig:sub2_lesion}
    \end{subfigure}
   % \hfill
    \begin{subfigure}{0.4\textwidth}
        \includegraphics[width=0.8\linewidth]{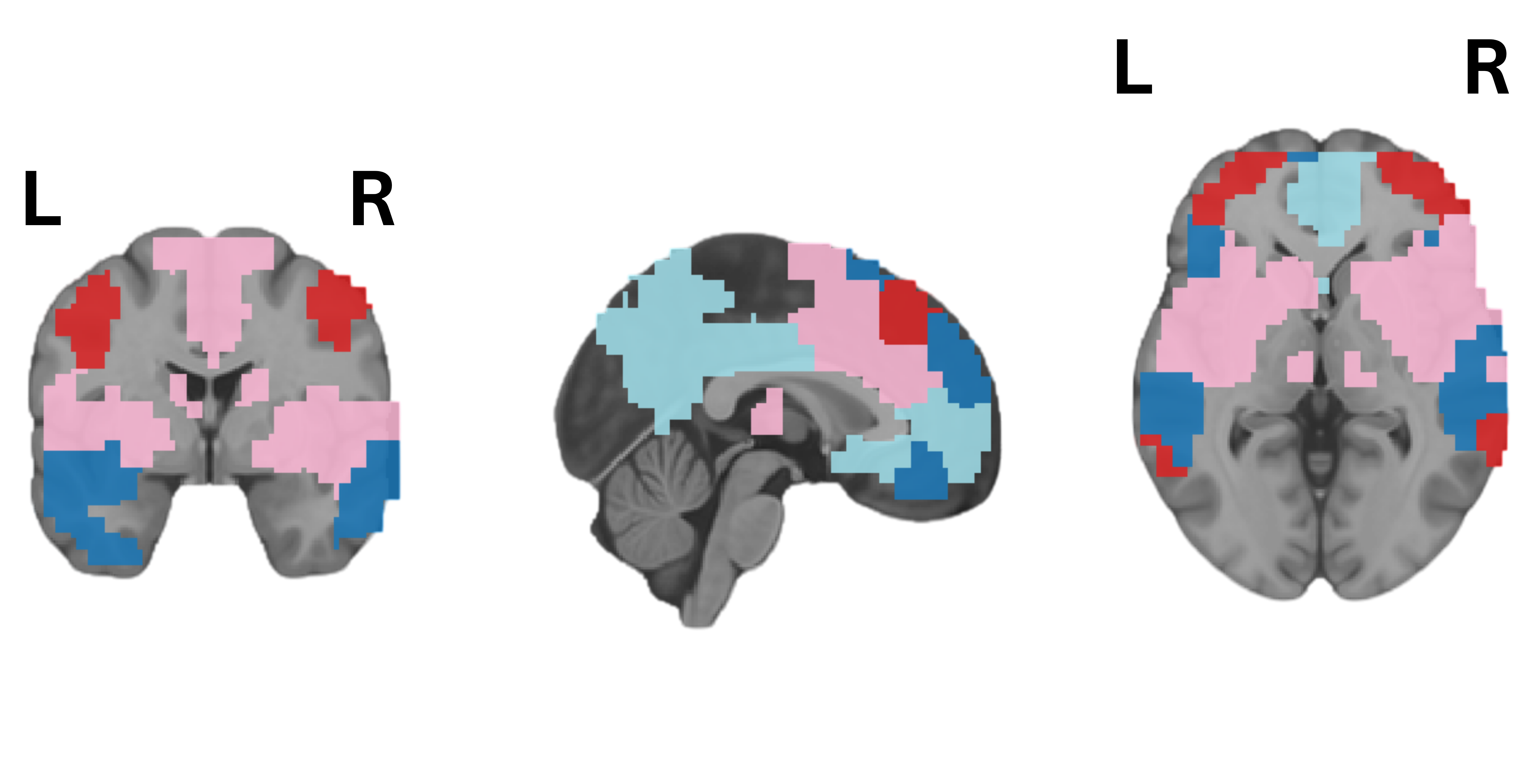}
        \caption{ABIDE}
        \label{fig:sub3_lesion}
    \end{subfigure}
    \begin{subfigure}{0.4\textwidth}
        \includegraphics[width=0.8\linewidth]{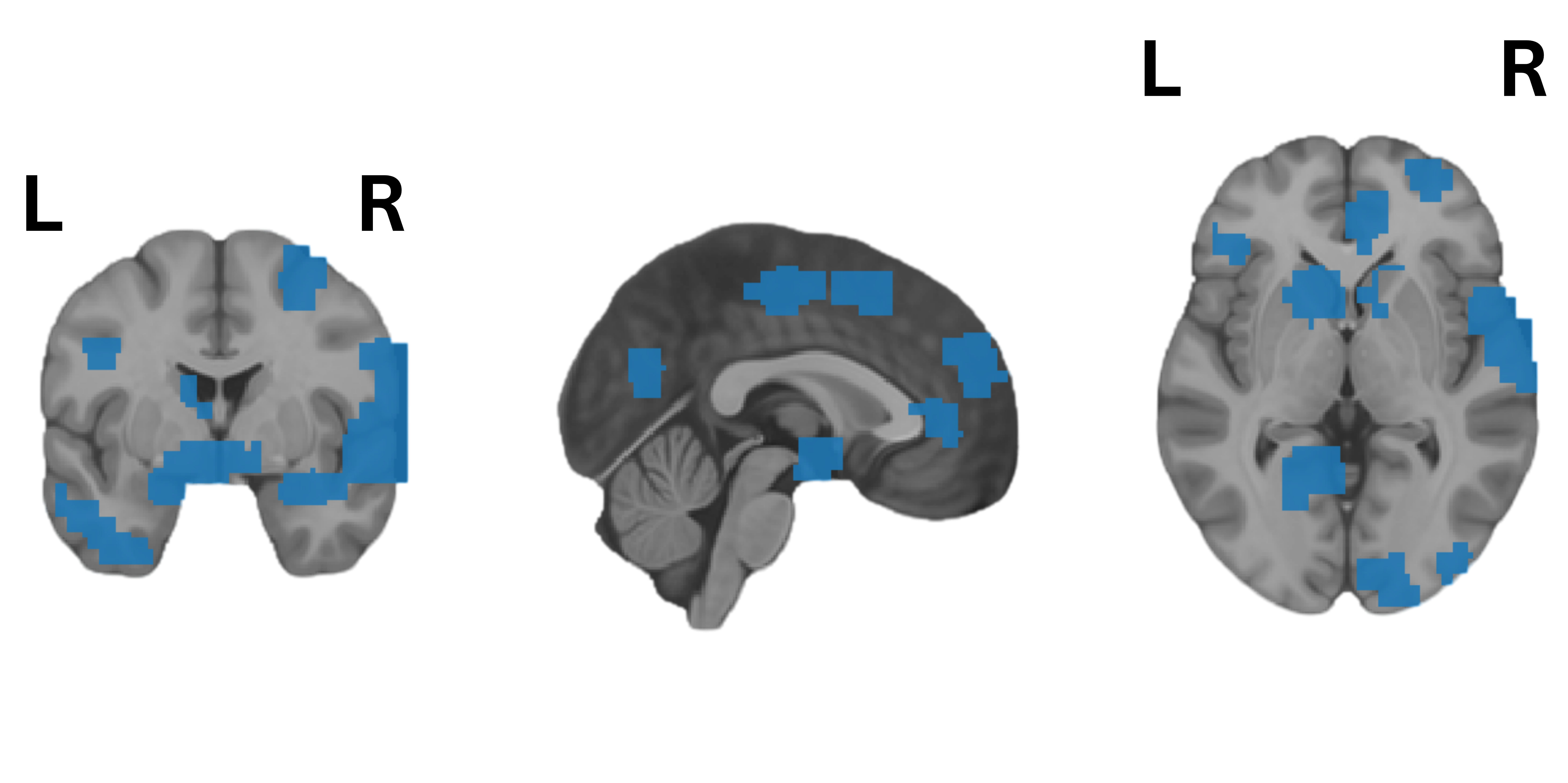}
        \caption{ADHD-200}
        \label{fig:sub4_lesion}
    \end{subfigure}
    \caption{Lesion visualization for four diseases. Colored regions indicate highly disease-relevant regions.}
    %\vspace{-1\baselineskip} 
    \label{fig:patho_vis}
\end{figure*}

To evaluate the effectiveness of our pathological pattern filter, Figure~\ref{fig:AD_patho_identify} presents the patho-scores obtained to identify PathoGraph for four different diseases. As illustrated, only a few of subgraphs yield patho-scores higher than the classification score $\alpha$. For Alzheimer’s disease, the PathoGraph is formed by remaining only the `Limbic' and `Subcortical' subgraphs (Figure~\ref{fig:sub1_patho}), in line with prior neuroscience findings~\citep{nelson2019limbic,Yang2021Nature}. For Parkinson's disease,  the `Cognitive Control', `DMN' (default mode network), and `Subcortical' subgraphs are identified as highly disease-relevant and remained to form the PathoGraph (Figure~\ref{fig:sub2_patho}), a finding highly consistent with existing neuroscience research~\citep{LuShared2024}. Given the absence of predefined region-of-interest (ROI) labels in the ABIDE and ADHD-200 brain graphs, we derived the PathoGraphs by identifying unlabeled subgraphs that yielded high patho-scores for the target disorders, namely Autism Spectrum Disorder and ADHD (Figures~\ref{fig:sub3_patho} and~\ref{fig:sub4_patho}). Meanwhile, we visualize the lesions identified by the filter for four diseases (Figure~\ref{fig:patho_vis}). The results indicate that the filter effectively eliminates less relevant regions and retains highly disease-relevant regions, enabling lesion localization. In addition, to demonstrate BrainPoG's ability to reveal disease-specific alterations, we visualize and compare group-wise heatmaps of PathoGraph representations. The results are shown in Figure~\ref{fig:patho_vis_compare_ADNI} and Figure~\ref{fig:patho_vis_compare_PPMI} in Appendix~\ref{appendix_patho_vis}.

\subsection{Ablation Study}

\paragraph{Effectiveness of Main Modules.}
Table~\ref{Tab:ab1_result} summarizes the ablation results (ACC and number of parameter) for the two main modules. F1 and AUC results are shown in Table~\ref{Tab:ab1_result2}, and results of running time and memory usage are provided in Table~\ref{Tab:ab1_result3} (see Appendix~\ref{appendix_ab_results}). Removing pathological feature distillation causes large, consistent accuracy drops of 36.12\%, 39.23\%, 34.99\%, and 29.41\% on ADNI, PPMI, ABIDE, and ADHD-200, respectively, indicating that distillation preserves disease-specific features and prevents noisy node features from misleading the classifier. Removing pathological pattern filter while retaining distillation increases parameter counts and produces mixed effects on accuracy. %However, our method still outperforms the full-graph variant on ADNI and ABIDE and achieves comparable performance on PPMI and ADHD-200. 
Removing both modules produces the lowest accuracy despite larger model size, confirming that the two-stage approach filters out noise and enhances disease-relevant representations. To further validate the effectiveness of noise feature dropping and pathological feature augmentation in our pathological feature distillation module, we conduct an ablation study by individually removing each component. The results are shown in Table~\ref{Tab:ab1_result_noise_augmentation} (Appendix~\ref{abltion_pathofeature}).

\begin{table*}[ht]
	\centering
	\caption{ Model performance (ACC (\%)) and efficiency (parameter number) of BrainPoG and its variants on four datasets. Pattern filter and feature distillation indicate pathological pattern filter and pathological feature distillation module, respectively.}
	\label{Tab:ab1_result}
	\renewcommand\arraystretch{1}
    \setlength{\tabcolsep}{3pt} % Adjust column separation	
    \small
    \resizebox{\textwidth}{!}{
    \begin{tabular}{ccc|cc|cc|cc}
    \toprule
    \multirow{2}{*}{Method} & \multicolumn{2}{c|}{ADNI} &\multicolumn{2}{c|}{PPMI} &\multicolumn{2}{c|}{ABIDE} &\multicolumn{2}{c}{ADHD-200} \\\cmidrule{2-9}
    & ACC&Parameter No.& ACC&Parameter No.&ACC&Parameter No.&ACC&Parameter No.\\
    \midrule
    w/o Pattern Filter& 69.86$\pm$16.31&1.0M &82.38$\pm$2.59&253K & 65.20$\pm$17.42 &5.1M& \textbf{96.07$\pm$3.45}&2.2M
    \\
    w/o Feature Distillation & 47.19$\pm$3.16 &240K&43.67$\pm$1.02 &\textbf{11K} &58.17$\pm$4.70 &1.7M&63.62$\pm$3.73&160K
    \\
    w/o Pattern Filter \& Feature Distillation& 44.47$\pm$1.61&1.0M&41.07$\pm$5.49&259K&52.33$\pm$3.14&5.1M&64.95$\pm$6.63 &2.3M
      \\ 
     \midrule
     \textbf{BrainPoG} & \textbf{83.31$\pm$4.90} & \textbf{227K}&\textbf{82.90$\pm$1.62}&\textbf{11K} & \textbf{93.16$\pm$2.46} &\textbf{415K}&93.03$\pm$3.27&\textbf{140K}
     \\
    \bottomrule
    \end{tabular}}
\end{table*}

\paragraph{Comparison between Different Dimensionality Reduction Methods.}

\begin{figure*}[ht]
    \centering
    \begin{subfigure}{0.24\textwidth}
        \includegraphics[width=\linewidth]{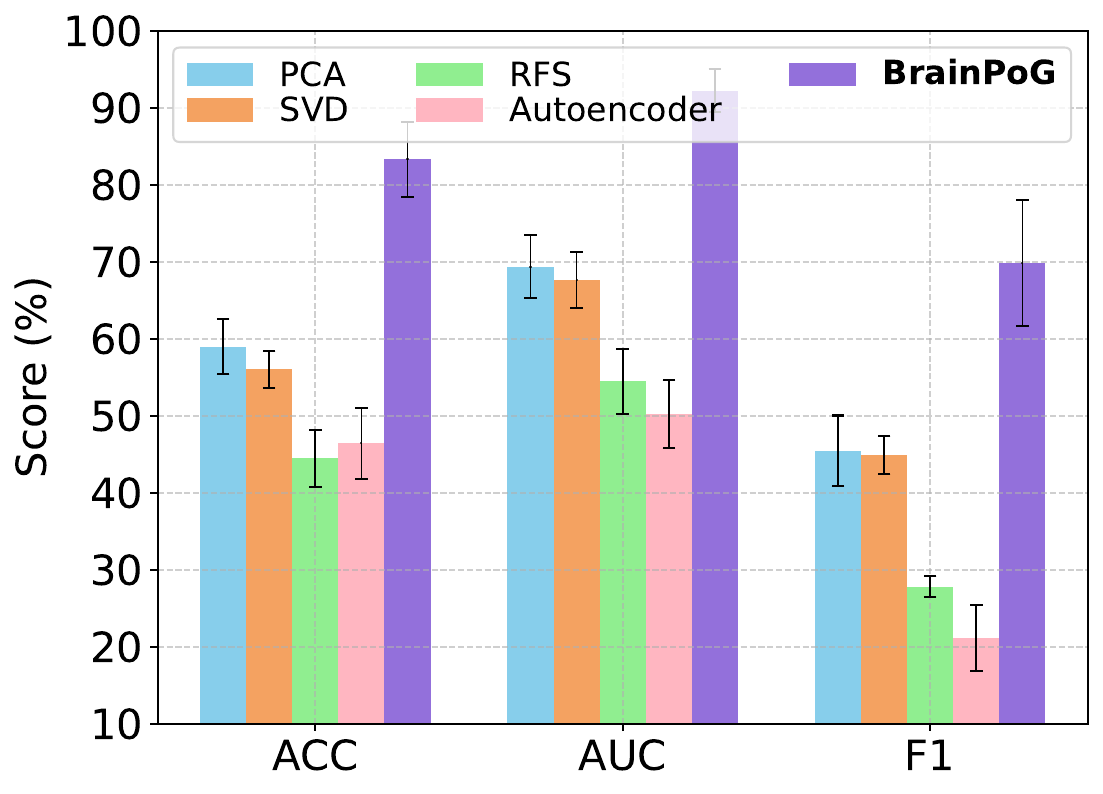}
        \caption{ADNI}
        \label{fig:sub1_ab2}
    \end{subfigure}
   % \hfill
    \begin{subfigure}{0.24\textwidth}
        \includegraphics[width=\linewidth]{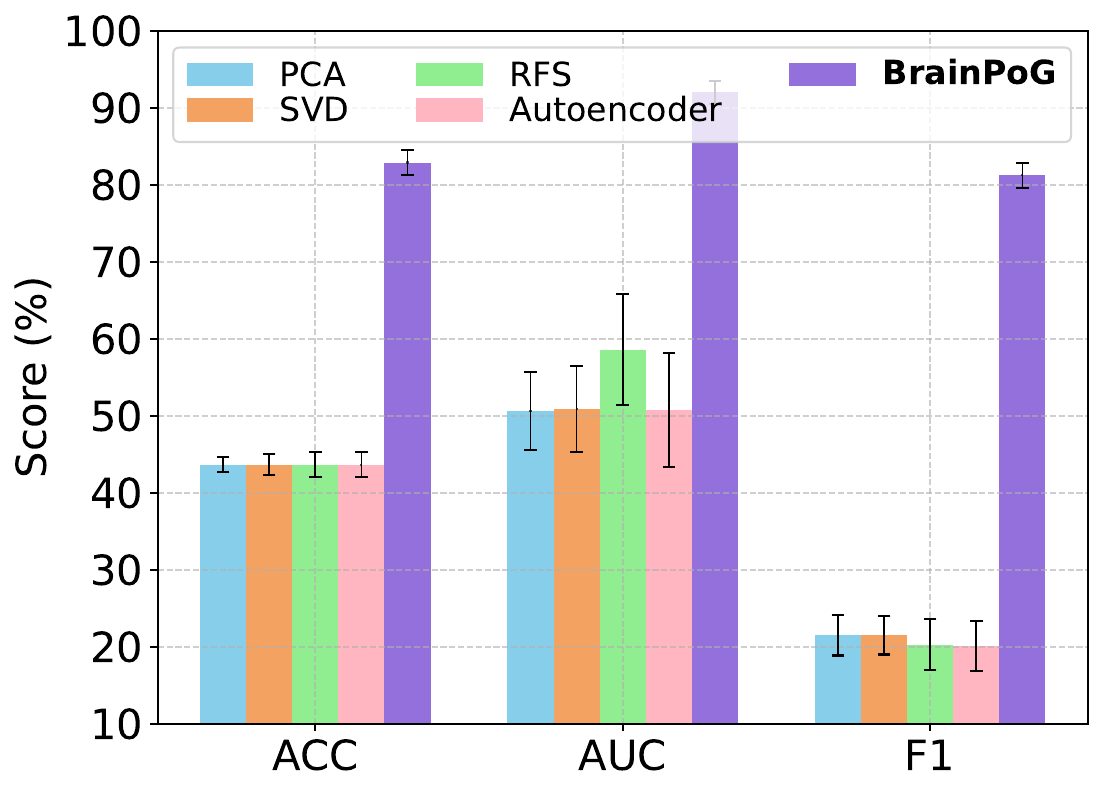}
        \caption{PPMI}
        \label{fig:sub2_ab2}
    \end{subfigure}
   % \hfill
    \begin{subfigure}{0.24\textwidth}
        \includegraphics[width=\linewidth]{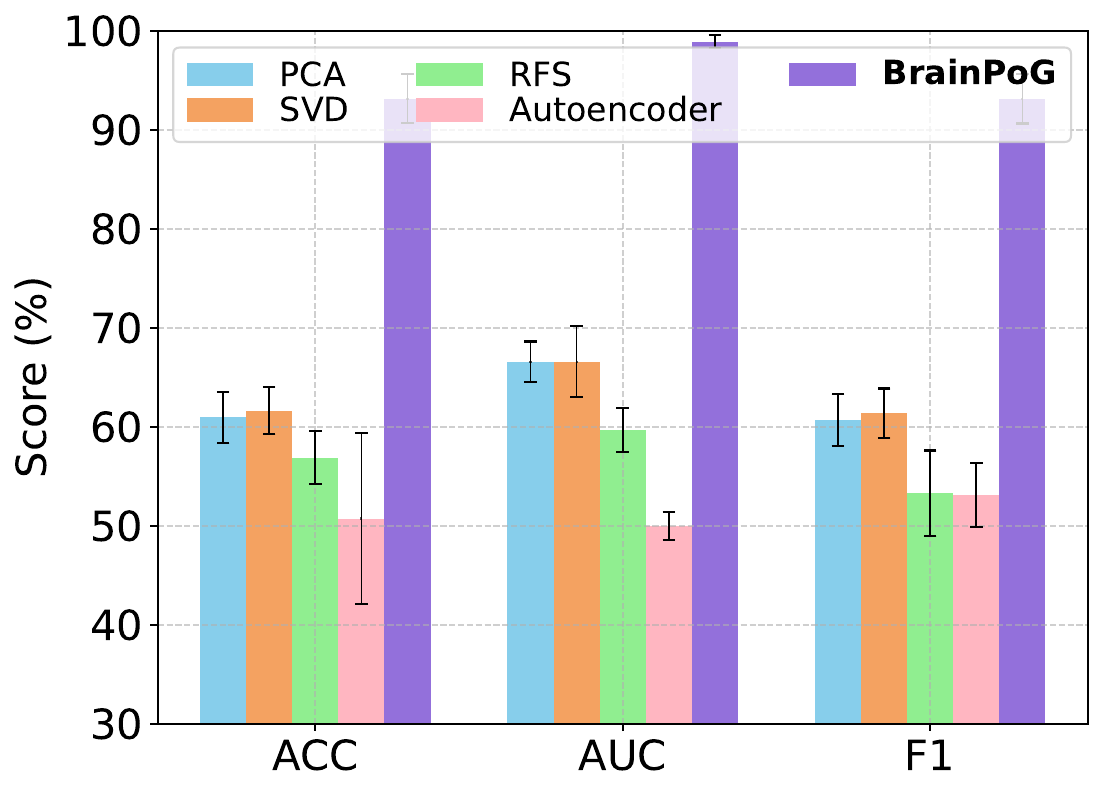}
        \caption{ABIDE}
        \label{fig:sub3_ab2}
    \end{subfigure}
    \begin{subfigure}{0.24\textwidth}
        \includegraphics[width=\linewidth]{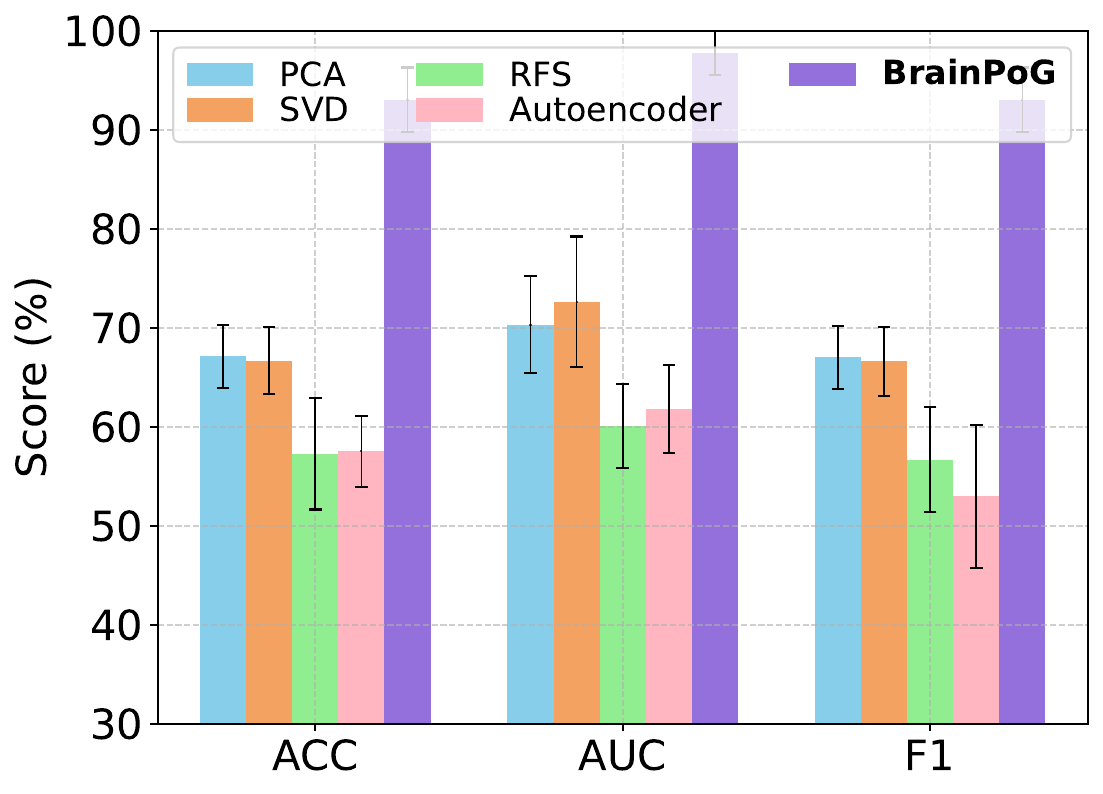}
        \caption{ADHD-200}
        \label{fig:sub4_ab2}
    \end{subfigure}
    \caption{Performance of models using different dimensionality reduction methods.}
   % \vspace{-1\baselineskip} 
    \label{fig:ab2_results}
\end{figure*}

To compare our pathological feature distillation method with other dimensionality reduction methods, we replace it with (1) PCA~\citep{WOLD198737pca}, (2) SVD~\citep{svd}, (3) Random feature selection (RFS)~\citep{10.5555/944919.944980rfs}, and (4) Autoencoder~\citep{bank2023autoencoders}, respectively. Figure~\ref{fig:ab2_results} compares the performance of models using different dimensionality reduction methods and demonstrates that BrianPoG consistently achieves better performance than other four methods across all four datasets. To assess computational overhead, Table~\ref{Tab:ab2_results} in Appendix~\ref{appendix_ab_results} reports parameter counts, per epoch running time and memory usage. Although the other methods demonstrate comparable model efficiency to BrainPoG, their performance deliver inferior performance.

\paragraph{Effectiveness of Community Detection Methods.}

\begin{wrapfigure}{r}{0.27\textwidth} 
    \centering
    %\vspace{-1.5\baselineskip}
    \includegraphics[width=0.27\textwidth]{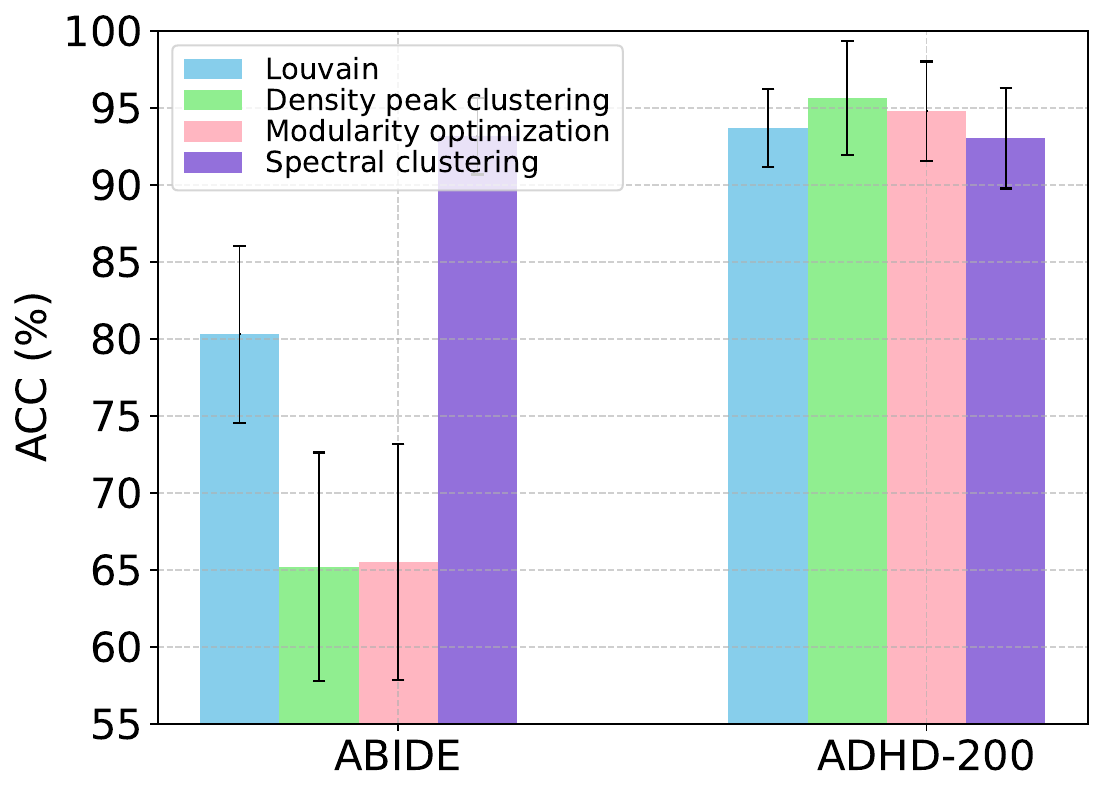}
    \captionsetup{width=0.27\textwidth} 
    \caption{Performance of models using different community detection methods.}
    \label{fig:ab3}
    \vspace{-3.5\baselineskip} 
\end{wrapfigure}

We use spectral clustering as the community detection method to generate subgraphs for the ABIDE and ADHD-200 datasets. To verify its effectiveness, we replace it with three alternatives, including Louvain, density peak clustering, and modularity optimization. As shown in Figure~\ref{fig:ab3}, model using spectral clustering yields the best performance on ABIDE and comparable results on ADHD-200. Notably, although it does not achieve the best performance on ADHD-200, its performance across two datasets is more stable compared to other methods.

\subsection{Hyperparameter Study}
 
\begin{figure*}[ht]
    \centering
    \vspace{-0.5\baselineskip} 
    \begin{subfigure}{0.24\textwidth}
        \includegraphics[width=\linewidth]{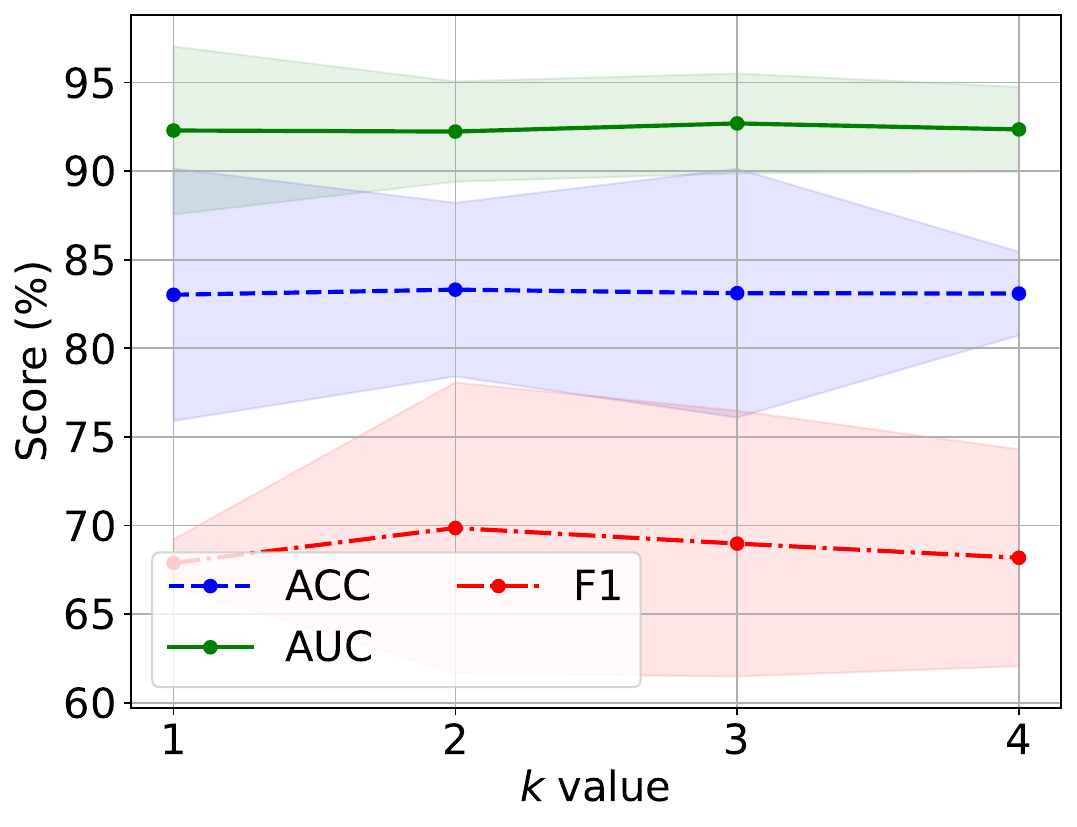}
        \caption{ADNI}
        \label{fig:sub1_hypara_k}
    \end{subfigure}
   % \hfill
    \begin{subfigure}{0.24\textwidth}
        \includegraphics[width=\linewidth]{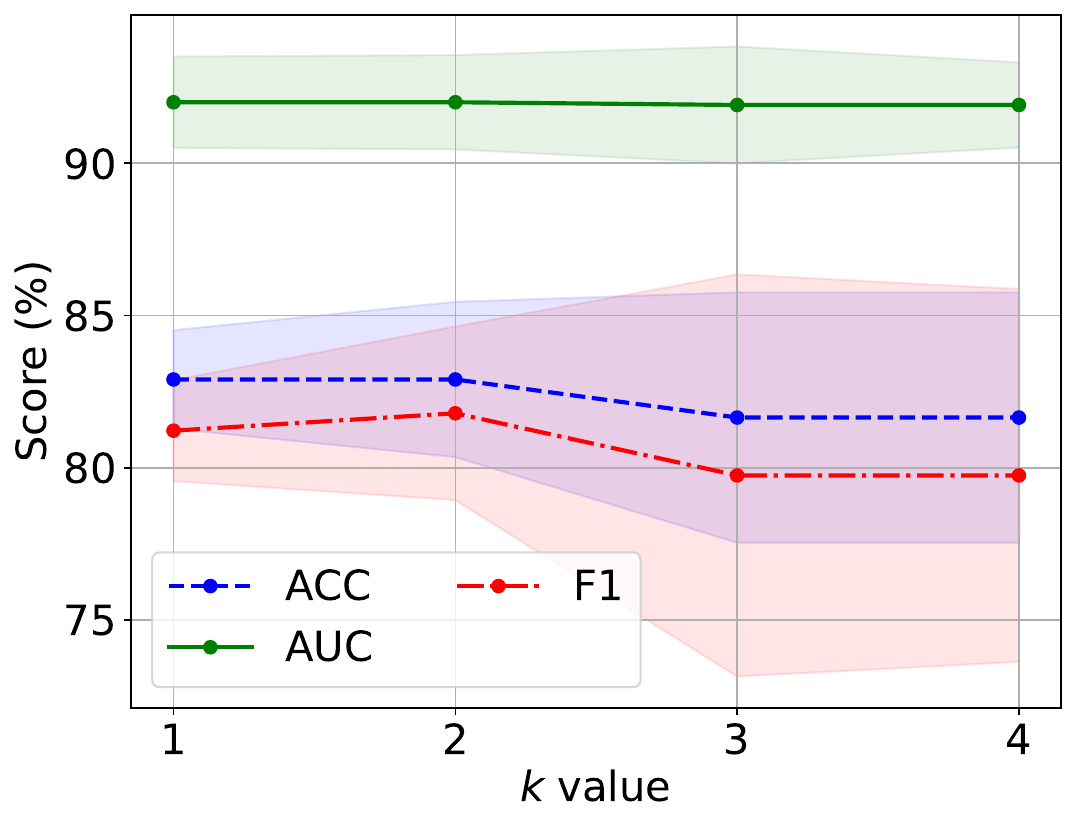}
        \caption{PPMI}
        \label{fig:sub2_hypara_k}
    \end{subfigure}
   % \hfill
    \begin{subfigure}{0.24\textwidth}
        \includegraphics[width=\linewidth]{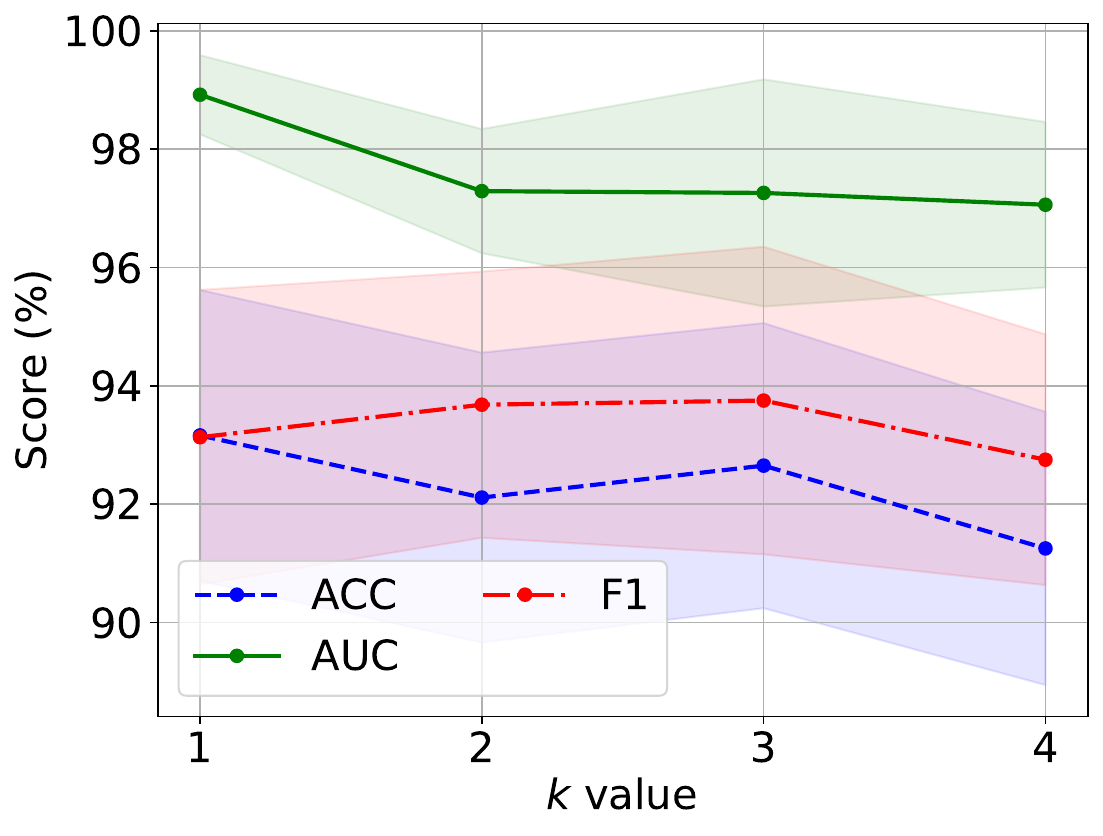}
        \caption{ABIDE}
        \label{fig:sub3_hypara_k}
    \end{subfigure}
    \begin{subfigure}{0.24\textwidth}
        \includegraphics[width=\linewidth]{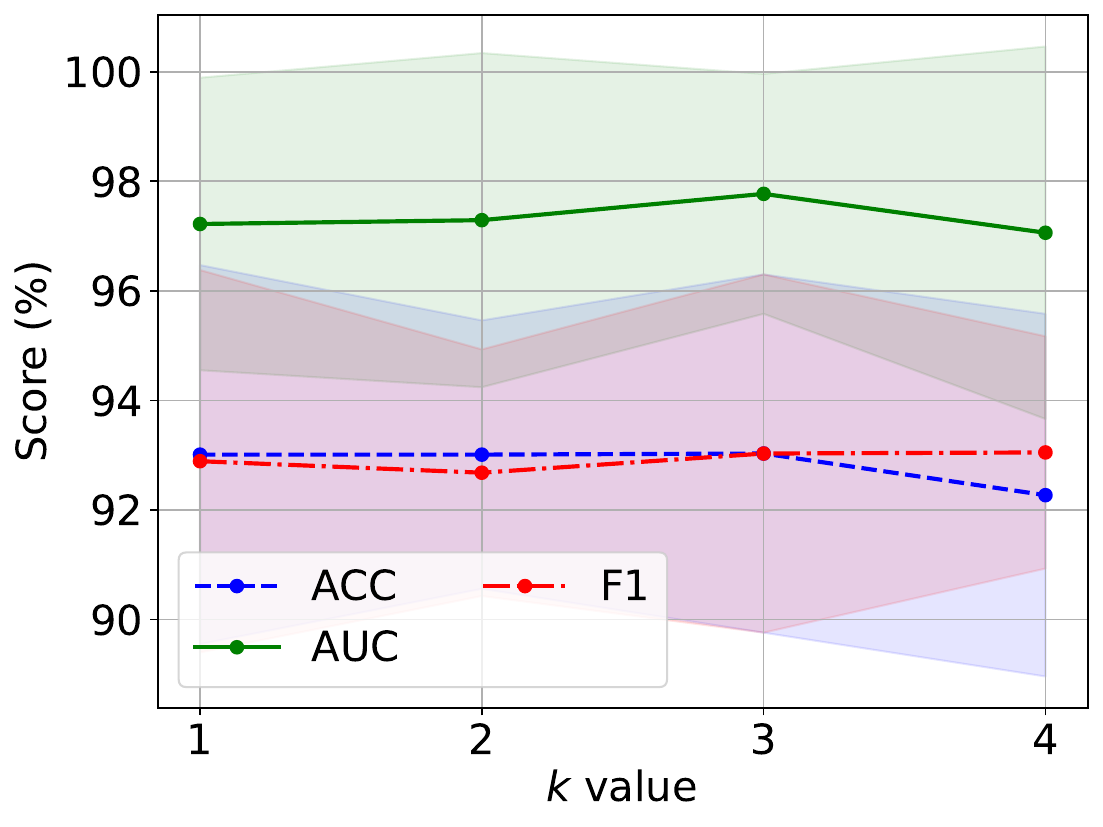}
        \caption{ADHD-200}
        \label{fig:sub4_hypara_k}
    \end{subfigure}
    \caption{Performance of models having different $k$ values.}
    \vspace{-1\baselineskip} 
    \label{fig:hyper1_results}
\end{figure*}

 \paragraph{Impact of $k$ Value.}
Noise feature dropping is performed by first computing the communal feature score matrix $\mathbf{S}$, followed by the removal of the top-$k$ and bottom-$k$ ranked features. To explore the impact of different $k$ values to model performance, we conduct experiments with setting $k$ to be 1, 2, 3, and 4. As shown in Figure~\ref{fig:hyper1_results}, the model performance remains stable across all $k$ values on all four datasets.

\paragraph{Impact of the Magnitude of Feature Enhancement ($\rho$).}

\begin{figure*}[ht]
    \centering
    \vspace{-0.5\baselineskip} 
    \begin{subfigure}{0.24\textwidth}
        \includegraphics[width=\linewidth]{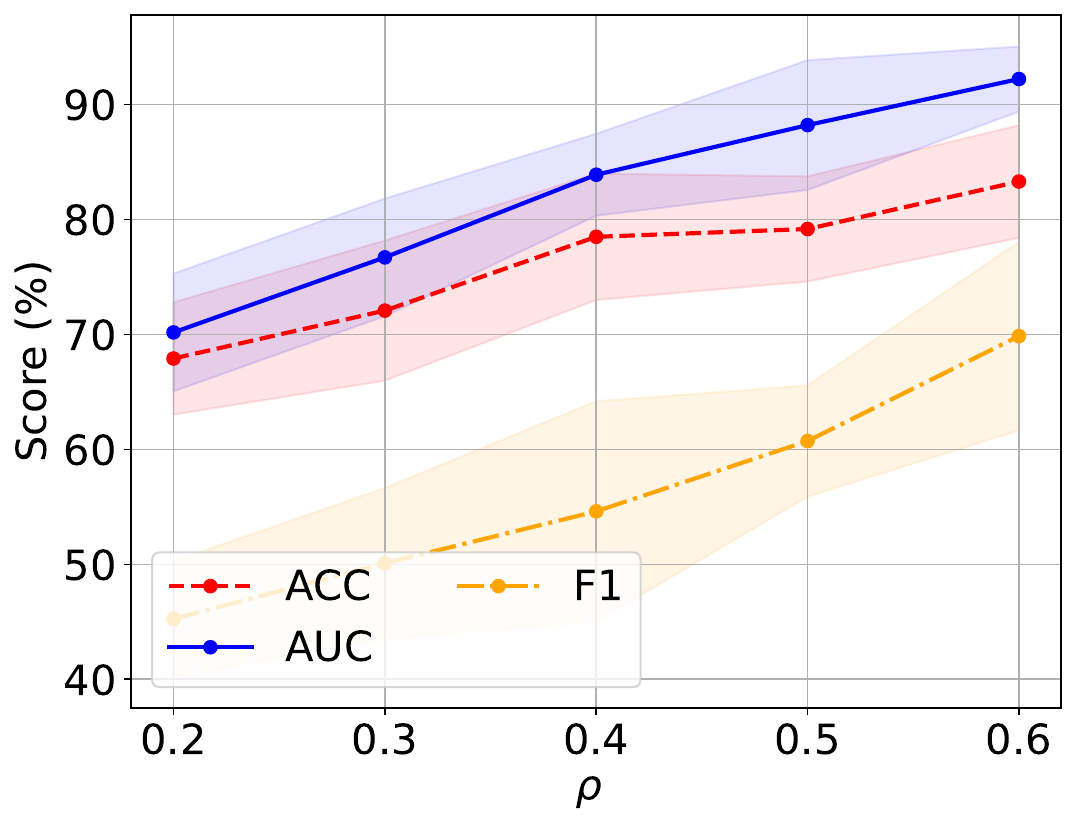}
        \caption{ADNI}
        \label{fig:sub1_hypara_rho}
    \end{subfigure}
   % \hfill
    \begin{subfigure}{0.24\textwidth}
        \includegraphics[width=\linewidth]{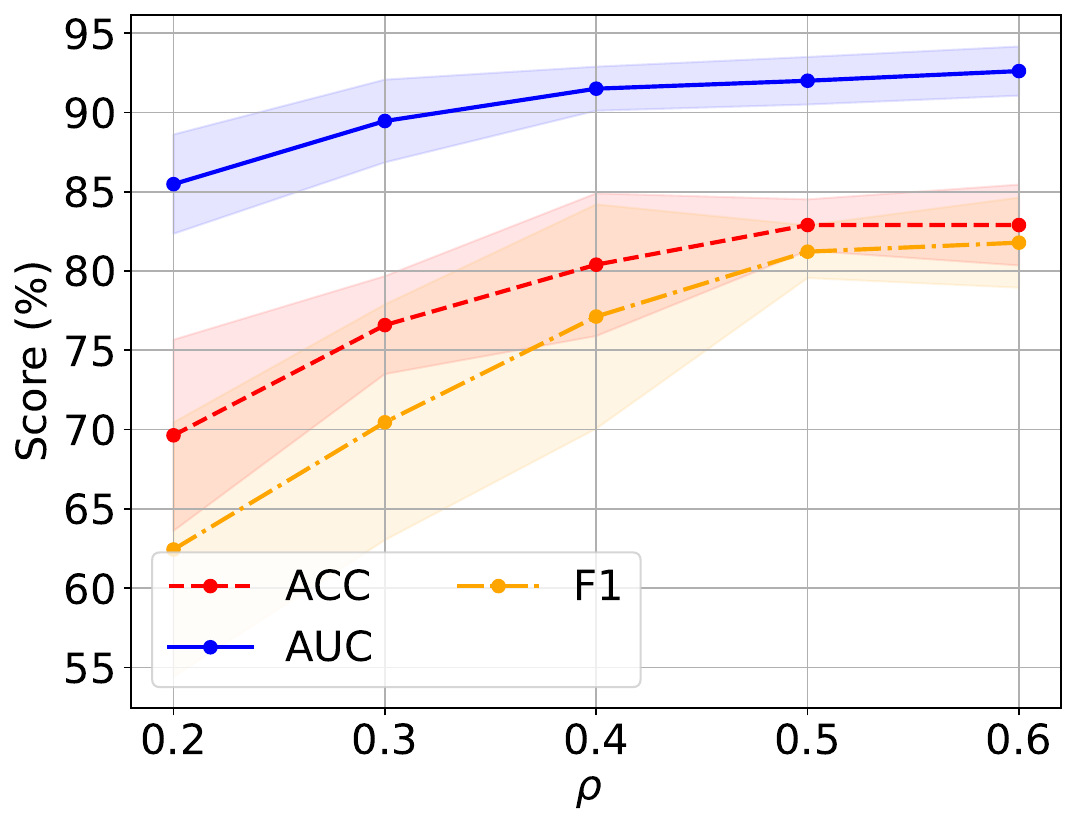}
        \caption{PPMI}
        \label{fig:sub2_hypara_rho}
    \end{subfigure}
   % \hfill
    \begin{subfigure}{0.24\textwidth}
        \includegraphics[width=\linewidth]{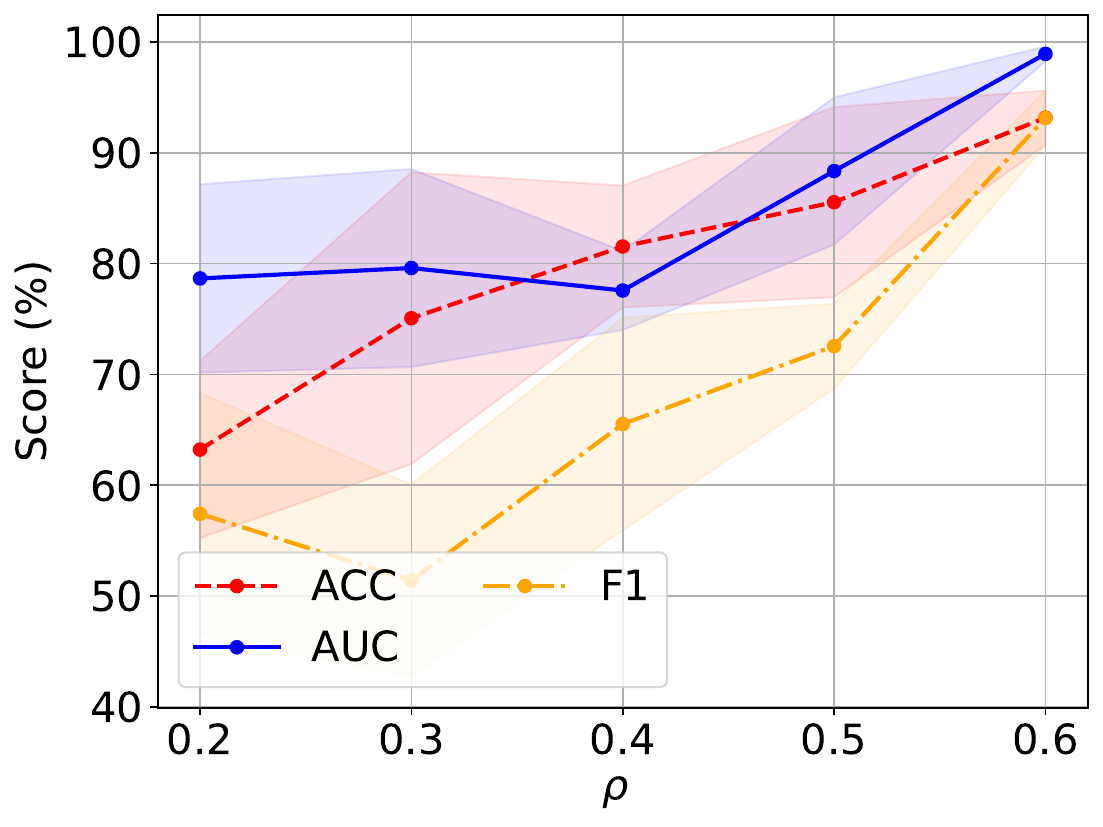}
        \caption{ABIDE}
        \label{fig:sub3_hypara_rho}
    \end{subfigure}
    \begin{subfigure}{0.24\textwidth}
        \includegraphics[width=\linewidth]{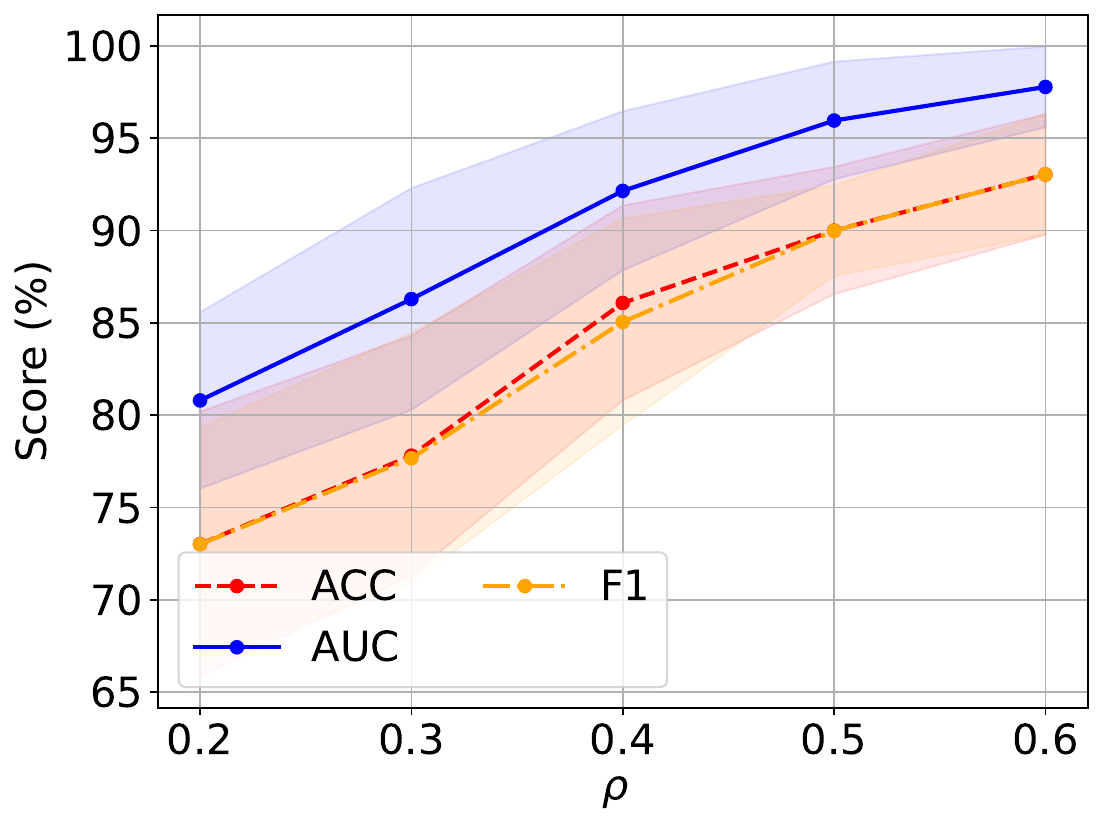}
        \caption{ADHD-200}
        \label{fig:sub4_hypara_rho}
    \end{subfigure}
    \caption{Model performance under different $\rho$ values.}
    \vspace{-1\baselineskip} 
    \label{fig:hyper2_results}
\end{figure*}

Figure~\ref{fig:hyper2_results} illustrates the model performance under different values of $\rho$. As the magnitude of feature enhancement increases, the model performance consistently improves across all four datasets. This is because stronger group-level feature enhancement makes it easier for the model to distinguish between different groups. These results highlight that our pathological feature augmentation method, particularly the group-specific feature importance guided augmentation strategy, plays a crucial role in enhancing the discriminative capability of the model for disease detection.

%\vspace{-0.5\baselineskip} 
\section{Conclusion}

This paper presents a lightweight BrainPoG model with a pathological pattern filter and a pathological feature distillation module. Significantly, BrainPoG can focus exclusively on disease-related knowledge while avoiding irrelevant information, enabling efficient brain graph learning. Extensive experiments on four benchmark datasets demonstrate BrainPoG exhibits superiority in both model performance and computational efficiency. This paper contributes to the intersection of neuroscience and artificial intelligence by advancing efficient brain graph learning and enhancing the practicality of models for real-world clinical applications. Although the results are encouraging, the sensitivity of BrainPoG to hyperparameters (e.g., GCN layer number and neuron size) remains a limitation. Therefore, it is worth exploring hyperparameter optimization and more robust architectures to further improve model stability.

\bibliography{iclr2026_conference}
\bibliographystyle{iclr2026_conference}

\newpage
\appendix
\section*{Appendix}

\section{Labels for AAL Atlas-based Brain Graphs}\label{appendix_roi_label}

\begin{table}[ht]
    \centering
    \caption{The labels of ROIs and subgraphs of brain graphs constructed based on AAL atlas.}
     \resizebox{0.8\textwidth}{!}{
    \begin{tabular}{c|c}        
        \toprule
        Subgraph & ROIs \\
        \midrule
        \multirow{1}{*}{Visual} & \begin{tabular}[t]{@{}l@{}}
        `Calcarine\_L', `Calcarine\_R', `Cuneus\_L', `Cuneus\_R',\\
        `Lingual\_L', `Lingual\_R', `Occipital\_Sup\_L', `Occipital\_Sup\_R',\\
        `Occipital\_Mid\_L', `Occipital\_Mid\_R', `Occipital\_Inf\_L', `Occipital\_Inf\_R'
        \end{tabular} \\
        \midrule
        
        \multirow{1}{*}{Auditory} & \begin{tabular}[t]{@{}l@{}}
        `Heschl\_L', `Heschl\_R', `Temporal\_Sup\_L', `Temporal\_Sup\_R',\\
        `Temporal\_Pole\_Sup\_L', `Temporal\_Pole\_Sup\_R'
        \end{tabular} \\
        \midrule
        
        \multirow{1}{*}{Cognitive Control} & \begin{tabular}[t]{@{}l@{}}
        `Frontal\_Sup\_L', `Frontal\_Sup\_R', `Cingulum\_Ant\_L', `Cingulum\_Ant\_R'
        \end{tabular} \\
        \midrule

        \multirow{1}{*}{Motor} & \begin{tabular}[t]{@{}l@{}}
        `Precentral\_L', `Precentral\_R', `Postcentral\_L', `Postcentral\_R',\\
        `Rolandic\_Oper\_L', `Rolandic\_Oper\_R', `Supp\_Motor\_Area\_L', `Supp\_Motor\_Area\_R'
        \end{tabular} \\
        \midrule

        \multirow{1}{*}{Limbic} & \begin{tabular}[t]{@{}l@{}}
        `Hippocampus\_L', `Hippocampus\_R', `Amygdala\_L', `Amygdala\_R',\\
        `ParaHippocampal\_L', `ParaHippocampal\_R',\\
        `Cingulum\_Mid\_L', `Cingulum\_Mid\_R', `Cingulum\_Post\_L', `Cingulum\_Post\_R'
        \end{tabular} \\
        \midrule

        \multirow{1}{*}{DMN} & \begin{tabular}[t]{@{}l@{}}
        `Frontal\_Sup\_Medial\_L', `Frontal\_Sup\_Medial\_R',\\
        `Frontal\_Med\_Orb\_L', `Frontal\_Med\_Orb\_R',\\
        `Precuneus\_L', `Precuneus\_R', `Angular\_L', `Angular\_R',\\
        `Cingulum\_Post\_L', `Cingulum\_Post\_R'
        \end{tabular} \\
        \midrule

        \multirow{1}{*}{Sensory} & \begin{tabular}[t]{@{}l@{}}
        `Postcentral\_L', `Postcentral\_R',\\
        `Parietal\_Sup\_L', `Parietal\_Sup\_R',\\
        `Parietal\_Inf\_L', `Parietal\_Inf\_R'
        \end{tabular} \\
        \midrule

        \multirow{1}{*}{Subcortical} & \begin{tabular}[t]{@{}l@{}}
        `Caudate\_L', `Caudate\_R', `Putamen\_L', `Putamen\_R',\\
        `Pallidum\_L', `Pallidum\_R', `Thalamus\_L', `Thalamus\_R'
        \end{tabular} \\
        \midrule

        \multirow{1}{*}{Olfactory} & \begin{tabular}[t]{@{}l@{}}
        `Olfactory\_L', `Olfactory\_R'
        \end{tabular} \\
        \bottomrule
    \end{tabular}}
    \label{tab:sub}
\end{table}

\section{Procedure}\label{produdure-appendix}

\begin{algorithm}
	\caption{Pathological Pattern Filter}
	\label{algo1}
	\begin{algorithmic}[1]
		\STATE \textbf{Input:} Brain graph dataset $\mathcal{B} = \{\mathcal{G}^{(1)}, \mathcal{G}^{(2)}, ..., \mathcal{G}^{(D)}\}$, where each $\mathcal{G}^{(d)}$ is divided into $m$ subgraphs $\{\mathcal{G}_1^{(d)}, \mathcal{G}_2^{(d)}, ..., \mathcal{G}_m^{(d)}\}$, $D$ is the number of samples.
		\STATE \textbf{Output:} PathoGraph set $\hat{\mathcal{B}}$=$\{\hat{\mathcal{G}}^{(1)}, \hat{\mathcal{G}}^{(2)}, ..., \hat{\mathcal{G}}^{(D)}\}$, where PathoGraph $\hat{\mathcal{G}}^{(d)}=\mathcal{G}^{(d)}-\{\mathcal{G}_i^{(d)}\}^{\hat{m}}_{i=1}$, $\hat{m}$ is the number of dropped subgraphs.
		
		\STATE Train the SVM with RBF kernel on complete graphs in $\mathcal{B}$. 
		\STATE Obtain classification score $\alpha$ for complete graphs in $\mathcal{B}$.
        \STATE Let $\hat{\mathcal{G}}=\mathcal{G}$.
		\FOR{subgraph $\mathcal{G}_i$ in $\mathcal{G}$}
            \STATE Construct subgraph dataset $\mathcal{B}_i=\{\mathcal{G}_i^{(1)},\mathcal{G}_i^{(2)},...,\mathcal{G}_i^{(D)}\}$.
			\STATE Input $\mathcal{B}_i$ into the SVM and obtain patho-score $\beta_i$.
			\IF{$\beta_i < \alpha$}
			\STATE $\hat{\mathcal{G}}=\hat{\mathcal{G}}-\mathcal{G}_i$.
		\ENDIF
		\ENDFOR		
		\STATE \textbf{Return} $\hat{\mathcal{B}}=\{\hat{\mathcal{G}}^{(1)}, \hat{\mathcal{G}}^{(2)}, ..., \hat{\mathcal{G}}^{(D)}\}$
	\end{algorithmic}
\end{algorithm}

\begin{algorithm}
\caption{Pathological Feature Distillation}
\label{algo2}
\begin{algorithmic}[1]
    \STATE \textbf{Input:} PathoGraph dataset $\hat{\mathcal{B}}=\{\hat{\mathbf{X}}^{(d)}\in\mathbb{R}^{\hat N\times \hat N}\}_{d=1}^{D}$, which can be divided into $Y$ groups.
    \STATE \textbf{Output:} Distilled and enhanced node representations $\{\widetilde{\mathbf{X}}^{\prime\,(d)}\}_{d=1}^{D}$.

    \STATE \textbf{Step 1: Noise Feature Dropping}
    \FOR{$i=1$ to $\hat N$}
        \STATE Construct cross-subject feature matrix $\mathbf{C}_{v_i}\in\mathbb{R}^{\hat N\times D}$ by stacking node $v_i$ across all $D$ subjects.
        \STATE Compute SVD: $\mathbf{C}_{v_i}=\mathbf{L}_{v_i}\mathbf{\Sigma}_{v_i}\mathbf{R}_{v_i}^{\top}$.
        \STATE Let $\mathbf{l}_{v_i}=\mathbf{L}_{v_i}[:,1]=[\mu_{v_i}^1,\dots,\mu_{v_i}^{\hat N}]^{\top}$.
        \STATE Update communal feature scoring matrix $\mathbf{S}[i,:]\leftarrow \mathbf{l}_{v_i}^{\top}$.
    \ENDFOR
    \FOR{$i=1$ to $\hat N$}
        \STATE Obtain top-$k$ and bottom-$k$ features for node $v_i$: $(|\mathbf{S}[i,:]|,\text{top-}{k})$; $(|\mathbf{S}[i,:]|,\text{bottom-}{k})$.       
        \STATE Drop top-$k$ and bottom-$k$ features.
    \ENDFOR
    \STATE Result: $\widetilde{\mathbf{X}}^{(d)} \in \mathbb{R}^{\hat N\times \hat N'}$, where $\hat N'=\hat N-2k$.

    \STATE \textbf{Step 2: Pathological Feature Augmentation}
    \FOR{$y=1$ to $Y$}
        \STATE Let $D_y=\{d\mid y^{(d)}=y\}$.
        \FOR{$i=1$ to $\hat N$}
            \STATE Construct group-wise matrix $\mathbf{C}^{(y)}_{v_i}\in\mathbb{R}^{\hat N'\times |D_y|}$ from $\{\widetilde{\mathbf{x}}^{(d)}_i\}_{d\in D_y}$.
            \STATE Compute SVD: $\mathbf{C}^{(y)}_{v_i}=\mathbf{L}^{(y)}_{v_i}\mathbf{\Sigma}^{(y)}_{v_i}\mathbf{R}^{(y)\top}_{v_i}$.
            \STATE Get group-specific feature scoring vector $\mathbf{F}_{y,i}\leftarrow \mathbf{L}^{(y)}_{v_i}[:,1]$.
        \ENDFOR
        \STATE Assemble $\mathbf{F}_{y}\in\mathbb{R}^{\hat N\times \hat N'}$ with rows $\mathbf{F}_{y,i}^{\top}$.
        \STATE Compute feature weights $\mathbf{W}\leftarrow \frac{1}{\hat N}\sum_{i=1}^{\hat N}|\widetilde{\mathbf{x}}^{(d)}_i|\odot \mathbf{F}_{y,i}$ for any $d\in D_y$.
        \FOR{$j=1$ to $\hat N'$}
            \STATE $p_j \leftarrow \min\!\left(\rho\cdot\frac{\log w_{max}-\log w_j}{\log w_{max}-\log \lambda_w},\,p_t\right)$.
            \STATE Sample $\tilde b_j \sim \text{Bernoulli}(1-p_j)$.
        \ENDFOR        
        \STATE Form group mask $\tilde{\mathbf{b}}\in\{0,1\}^{\hat N'}$.
        \FORALL{$d \in D_y$}
            \FOR{$i=1$ to $\hat N$}
                \STATE $\widetilde{\mathbf{x}}^{\prime\,(d)}_i \leftarrow \widetilde{\mathbf{x}}^{(d)}_i \odot \tilde{\mathbf{b}}$.
            \ENDFOR
            \STATE Stack rows to obtain $\widetilde{\mathbf{X}}^{\prime\,(d)}$.
        \ENDFOR
    \ENDFOR
    \STATE \textbf{Return:} $\{\widetilde{\mathbf{X}}^{\prime\,(d)}\}_{d=1}^{D}$.
\end{algorithmic}
\end{algorithm}

\newpage

\section{Proof of Proposition 1}\label{proof1}

\textbf{Proposition~\ref{proposition1}} (Communal features).~
Based on feature scores, the top ranked features in the cross-subject feature representation $\mathbf{C}_{v_i}$ are considered as communal features of node $v_i$ across all subjects.

\begin{proof}
For the cross-subject feature matrix $\mathbf{C}_{v_i}\in\mathbb{R}^{\hat N\times D}$ of node $v_i$, its SVD is
\[
\mathbf{C}_{v_i}=\mathbf{L}_{v_i}\mathbf{\Sigma}_{v_i}\mathbf{R}_{v_i}^{\top},\quad 
\mathbf{\Sigma}_{v_i}=\operatorname{diag}(\sigma_1,\sigma_2,\ldots),\ \sigma_1\ge\sigma_2\ge\cdots\ge 0.
\]
Let the first left singular vector be
\[
\mathbf{l}_{v_i}=\mathbf{L}_{v_i}[:,1]=
\begin{bmatrix}
\mu^1_{v_i},\mu^2_{v_i},\ldots,\mu^{\hat N}_{v_i}
\end{bmatrix}^{\top}.
\]

By the variational characterization of singular values,
\[
\sigma_1 \;=\; \max_{\|\mathbf{w}\|_2=\|\mathbf{z}\|_2=1}\ \mathbf{w}^{\top}\mathbf{C}_{v_i}\mathbf{z},
\]
with the maximizer attained at $\mathbf{w}=\mathbf{l}_{v_i}$ and $\mathbf{z}=\mathbf{r}_{v_i}$, where $\mathbf{r}_{v_i}$ is the first right singular vector. Therefore, $\mathbf{l}_{v_i}$ is the feature-space direction that captures the maximum across-subject shared variation.

Moreover, by the Eckart--Young theorem,
\[
\arg\min_{\operatorname{rank}(\mathbf{M})=1}\ \|\mathbf{C}_{v_i}-\mathbf{M}\|_F
\;=\; \sigma_1\,\mathbf{l}_{v_i}\mathbf{r}_{v_i}^{\top},
\]
where $\mathbf{M}$ ranges over all rank-one matrices. Hence, the best rank-one feature of $\mathbf{C}_{v_i}$ is given by $\sigma_1\,\mathbf{l}_{v_i}\mathbf{r}_{v_i}^{\top}$, and the contribution of each feature dimension to this feature is encoded in the entries of $\mathbf{l}_{v_i}$. Consequently, the score $\mu^j_{v_i}$ quantifies how strongly the $j$-th feature contributes to the dominant across-subject shared component. Ranking features by $\mu^j_{v_i}$ therefore identifies those with the highest shared contribution across subjects, which are considered as the communal features.

\end{proof}

\section{Implementation Details}\label{imple}

The detailed hyperparameter settings for training BrainPoG on four datasets are summarized in Table~\ref{para}. The model parameters are trained using the Adam optimizer. For the ADNI and PPMI datasets, subgraphs are divided according to the provided labels. Therefore, the community number is not applicable and is denoted as “/”.

 \begin{table}[!ht]
	\centering
	\caption{Hyperparameters for training on four different datasets.}
	\label{para}
	\renewcommand\arraystretch{1}
	\begin{tabular}{c|cccc}
		\toprule
		Hyperparameter& ADNI&PPMI&ABIDE&ADHD-200\\\midrule
        Noise dropping level ($k$)&2&1&1&3 \\
        Feature enhancement magnitude ($\rho$) & 0.6&0.5&0.6&0.6\\
        Community number&/&/&7&5\\
        \#Layers&4&2&2&2\\
        \#Neurons&128&32&128&64\\
        Dropout&0.6&0.6&0.6&0.6\\
        Learning rate&5e-3&5e-3&5e-3&5e-3\\
        \#Epochs&200&200&200&200\\
        Weight decay&3e-3&3e-3&3e-3&3e-3\\
		\bottomrule	
	\end{tabular}
\end{table}

\section{Additional Experimental Results}\label{add_results}

\subsection{Interpretability Analysis}\label{appendix_interpretability}
\begin{figure}[ht]
    \centering
    \includegraphics[width=0.85\linewidth]{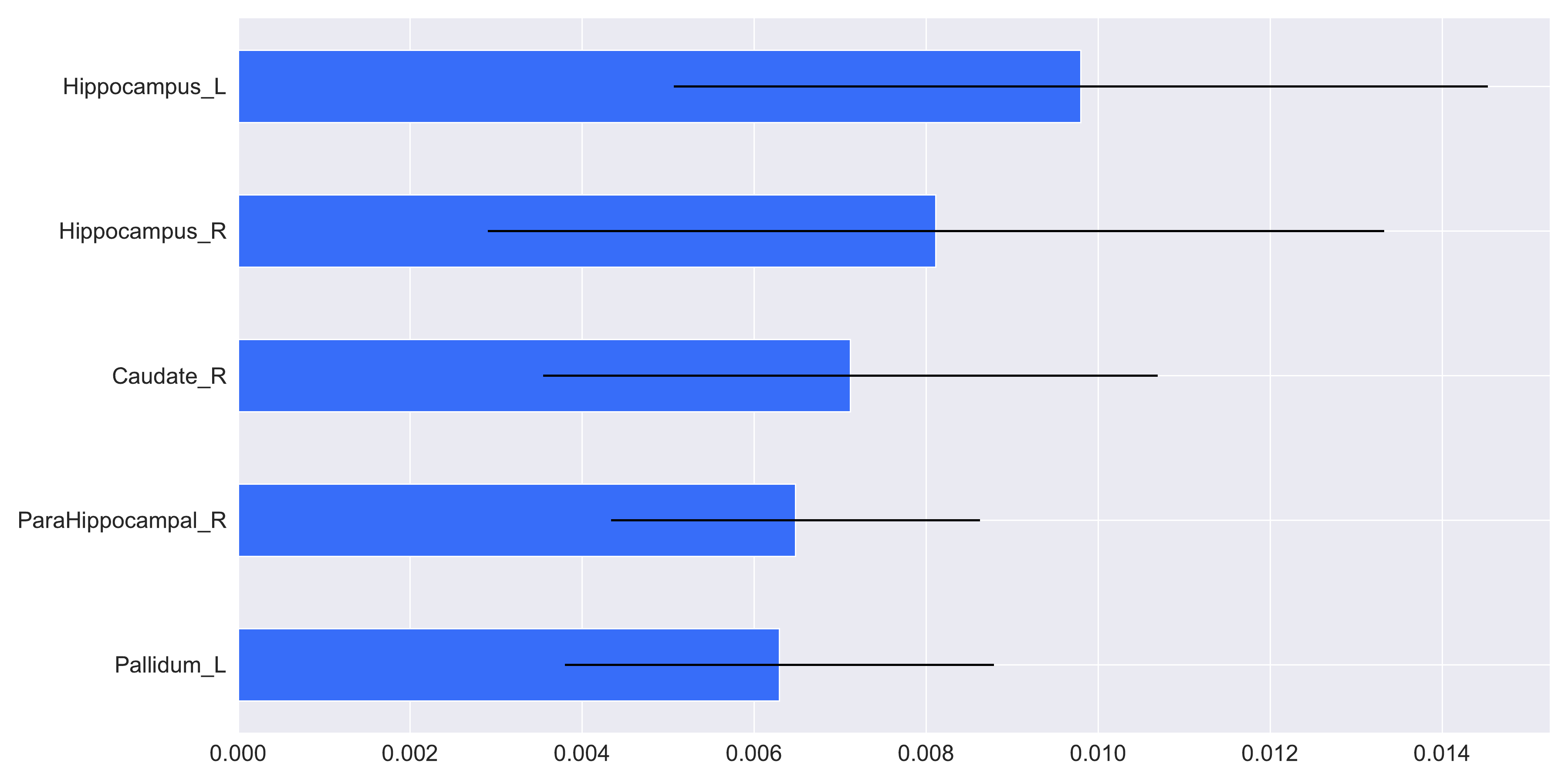}
    \caption{Visualization of the top-5 ROIs with the highest SHAP values for ADNI dataset.}
    \label{fig:shap_ADNI}
\end{figure}

\begin{figure}[ht]
    \centering
    \includegraphics[width=0.85\linewidth]{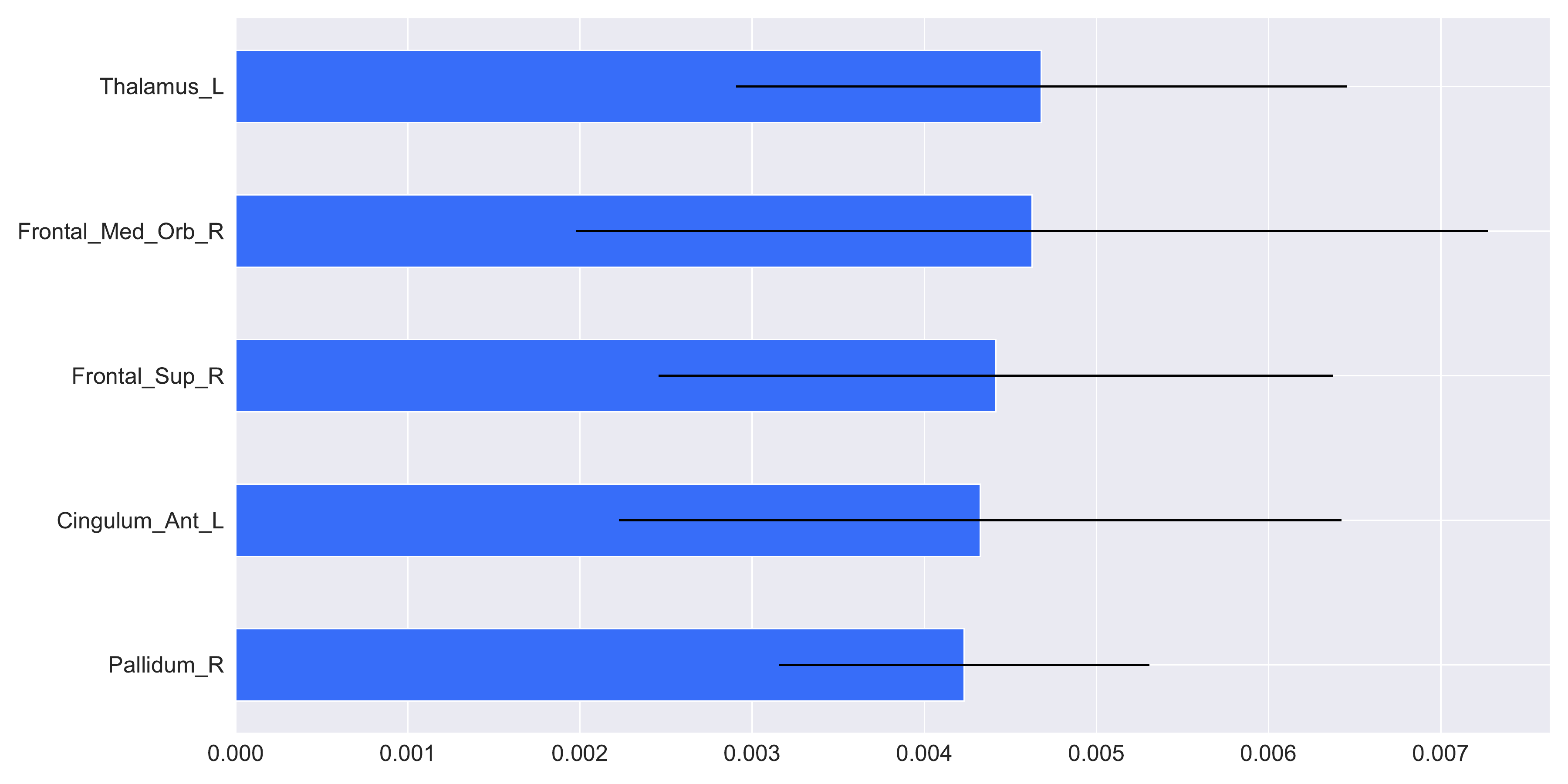}
    \caption{Visualization of the top-5 ROIs with the highest SHAP values for PPMI dataset.}
    \label{fig:shap_PPMI}
\end{figure}

We apply the SHAP model to evaluate the interpretability of the proposed method. By computing the SHAP values of the learned brain graph representations, we identified the brain ROIs most influential in brain disease.
Notably, since the ABIDE and ADHD-200 datasets do not provide ROI-level labels, the interpretability analysis is conducted only on the ADNI and PPMI datasets.
Figure~\ref{fig:shap_ADNI} illustrate the top-5 brain ROIs with the highest SHAP values identified in the ADNI dataset, including the hippocampus, caudate, parahippocampal, and pallidum.
These ROIs exhibit strong associations with ADNI-related diseases, in particular, pathological alterations in the hippocampus are directly implicated in the development of Alzheimer’s disease~\citep{Salta2023Adult}.
For Parkinson’s disease, the thalamus, frontal cortex, and cingulum are identified as the ROIs with the highest SHAP values, as illustrated in Figure~\ref{fig:shap_PPMI}.
The identified ROIs show strong associations with Parkinson’s disease predictions, which aligns closely with evidence reported in prior neuroscience studies on Parkinson’s disease~\citep{ChenReduced2023,Pagonabarraga2024}.
These results demonstrate that BrainPoG effectively identifies disease-relevant brain ROIs, providing interpretable insights that align with established neuroscience knowledge for both Alzheimer’s and Parkinson’s diseases.

\subsection{Time and Space Complexity}\label{appendix_efficiency}

To further verify the efficiency of BrainPoG, we compare its per-graph inference complexity with other state-of-the-art brain graph learning models (see Table~\ref{tab:complexity}). For all baselines, brain graphs are typically dense and require processing all ROIs, leading to both time and space complexity of order $\mathcal{O}(N^{2})$, with constant factors depending on the number of layers, heads, or additional modules. In contrast, BrainPoG performs inference on a filtered PathoGraph of retained subgraphs, which reduces the effective graph size to $\hat{N} \ll N$, resulting in both time and space complexity of $\mathcal{O}(\hat{N}^{2})$. Thus, while the asymptotic order remains quadratic, BrainPoG operates on a much smaller graph and achieves substantially lower running time and memory consumption during per-graph inference.

\begin{table}[ht]
    \centering
    \caption{Time and space complexity comparison between BrainPoG and state-of-the-art brain graph learning models.}
    \begin{tabular}{c|c|c}
    \toprule
         Method&  Time complexity&Space complexity\\\midrule
         BrainGB& $\mathcal{O}_t(N^2)$&$\mathcal{O}_s(N^2)$ \\ 
         BRAINNETTF& $\mathcal{O}_t(N^2)$&$\mathcal{O}_s(N^2)$ \\ 
         A-GCL& $\mathcal{O}_t(N^2)$&$\mathcal{O}_s(N^2)$ \\ 
         MCST-GCN& $\mathcal{O}_t(N^2)$&$\mathcal{O}_s(N^2)$ \\ 
         ALTER& $\mathcal{O}_t(N^2)$&$\mathcal{O}_s(N^2)$ \\ 
         BioBGT& $\mathcal{O}_t(N^2)$&$\mathcal{O}_s(N^2)$ \\ 
         BrainOOD& $\mathcal{O}_t(N^2)$&$\mathcal{O}_s(N^2)$ \\\midrule
         \textbf{BrainPoG}&$\mathcal{O}_t(\hat{N}^2)$&$\mathcal{O}_s(\hat{N}^2)$
         
         \\\bottomrule 
    \end{tabular}
    \label{tab:complexity}
\end{table}

\subsection{PathoGraph Visualization}\label{appendix_patho_vis}

\begin{figure*}[ht]
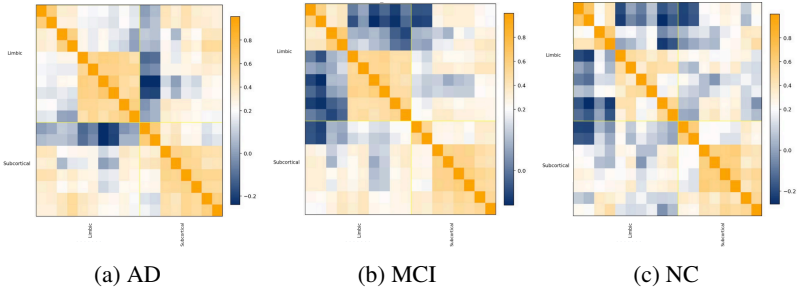

    \centering
    \begin{subfigure}{0.25\textwidth}
        \includegraphics[width=\linewidth]{fig/patho_vis_compare_ADNI_AD.pdf}
        \caption{AD}
        \label{fig:sub1_patho_vis_compare}
    \end{subfigure}
   % \hfill
    \begin{subfigure}{0.25\textwidth}
        \includegraphics[width=\linewidth]{fig/patho_vis_compare_ADNI_MCI.pdf}
        \caption{MCI}
        \label{fig:sub2_patho_vis_compare}
    \end{subfigure}
   % \hfill
    \begin{subfigure}{0.25\textwidth}
        \includegraphics[width=\linewidth]{fig/patho_vis_compare_ADNI_NC.pdf}
        \caption{NC}
        \label{fig:sub3_patho_vis_compare}
    \end{subfigure}
    \caption{Group-wise heatmaps of PathoGraph representations in ADNI dataset.}
    \vspace{-1\baselineskip} 
    \label{fig:patho_vis_compare_ADNI}
\end{figure*}

\begin{figure*}[ht]
    \centering
    \begin{subfigure}{0.25\textwidth}
        \includegraphics[width=\linewidth]{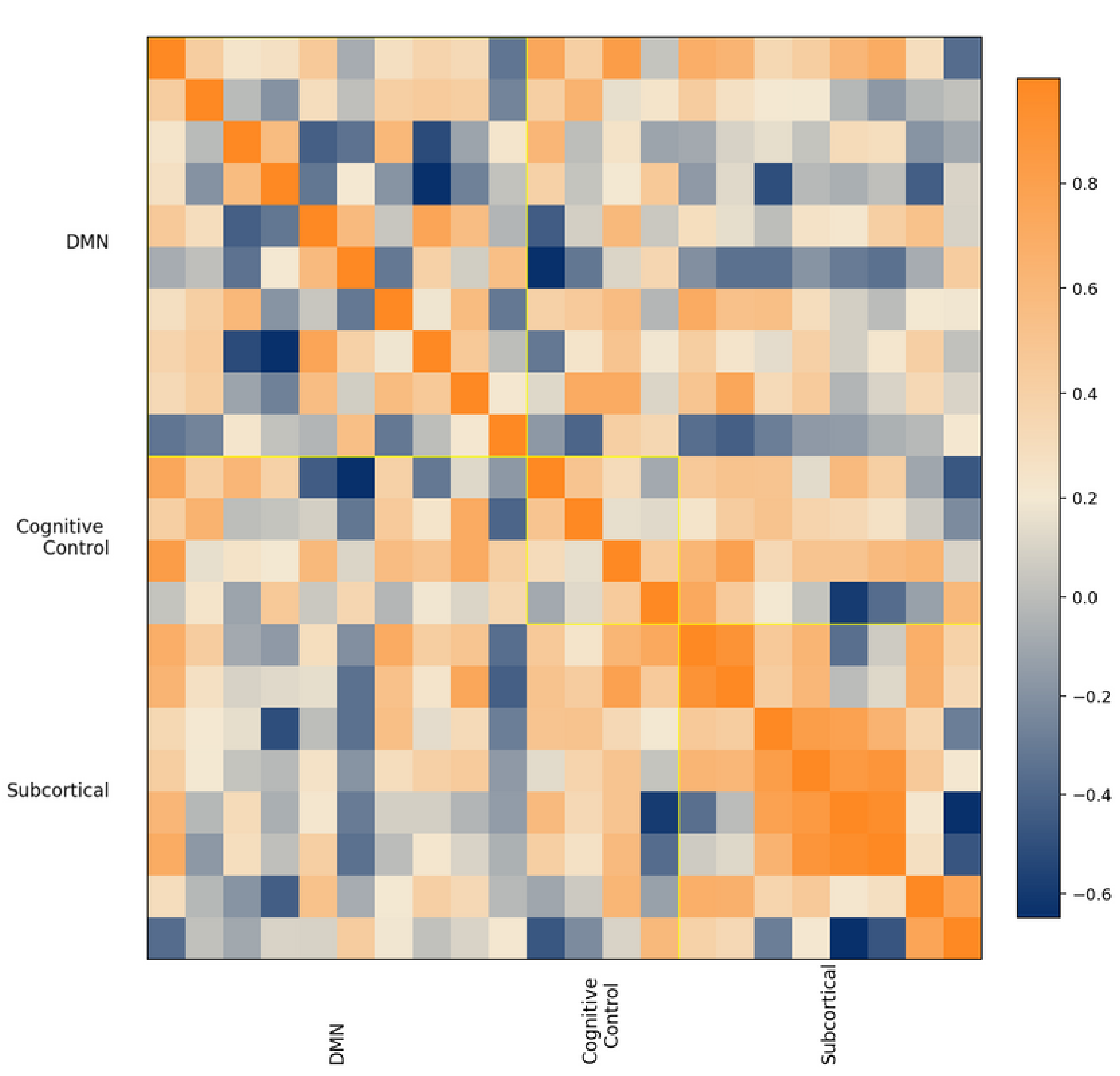}
        \caption{PD}
        \label{fig:sub1_patho_vis_compare_PPMI}
    \end{subfigure}
   % \hfill
    \begin{subfigure}{0.25\textwidth}
        \includegraphics[width=\linewidth]{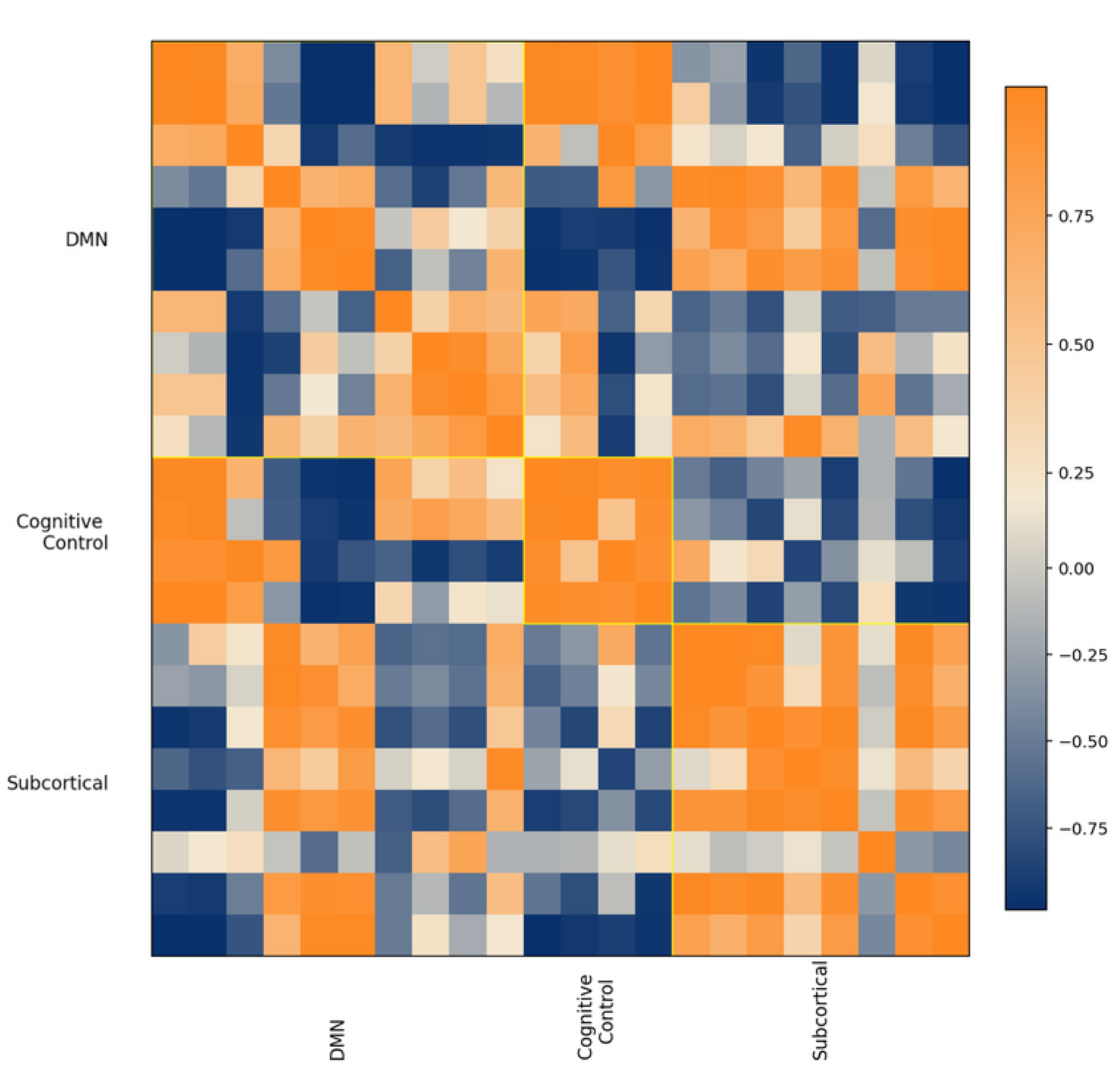}
        \caption{Prodromal}
        \label{fig:sub2_patho_vis_compare_PPMI}
    \end{subfigure}
   % \hfill
    \begin{subfigure}{0.25\textwidth}
        \includegraphics[width=\linewidth]{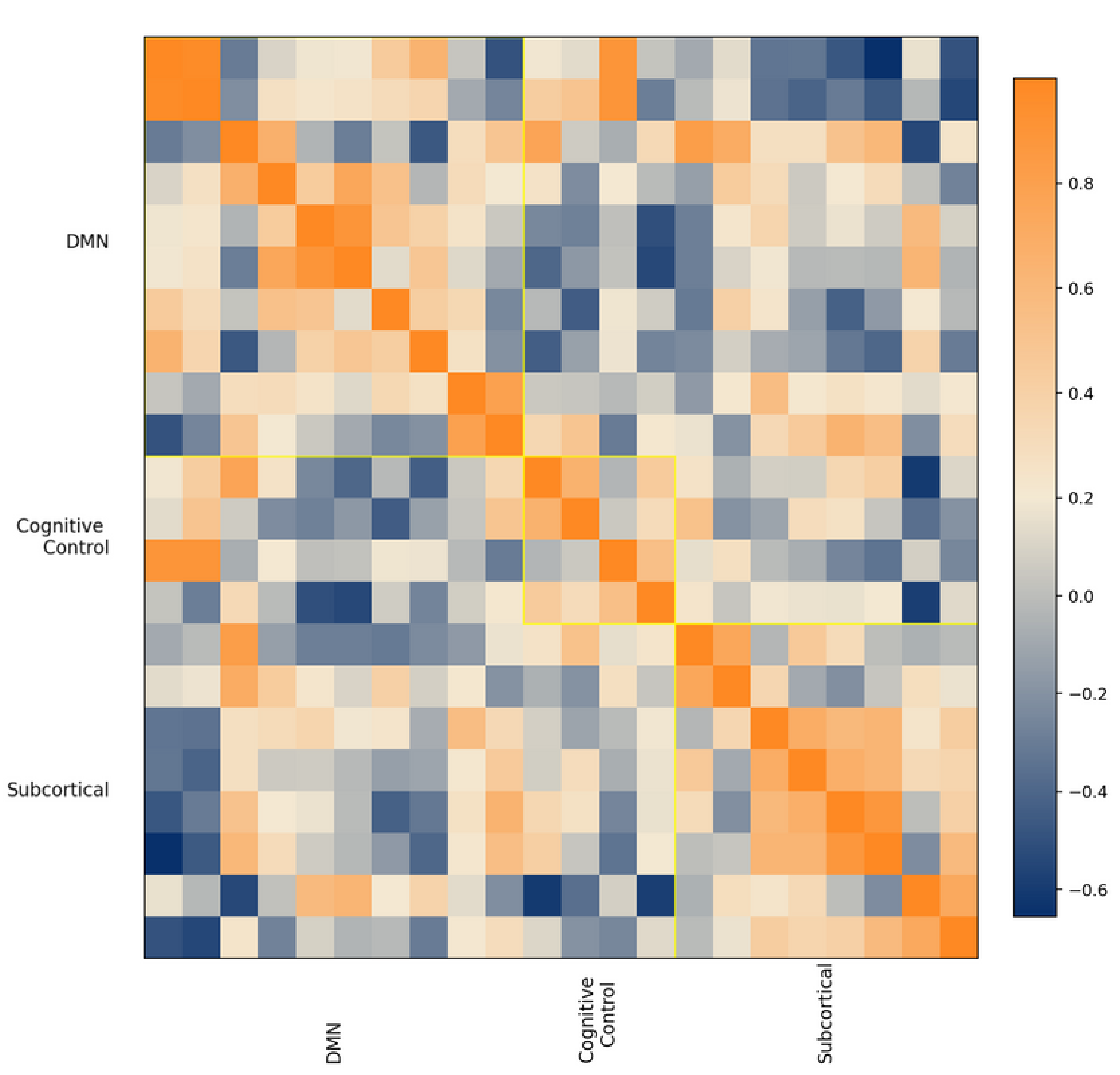}
        \caption{NC}
        \label{fig:sub3_patho_vis_compare_PPMI}
    \end{subfigure}
    \caption{Group-wise heatmaps of PathoGraph representations in PPMI dataset.}
    \label{fig:patho_vis_compare_PPMI}
\end{figure*}

Figures~\ref{fig:patho_vis_compare_ADNI} and \ref{fig:patho_vis_compare_PPMI} present group-wise heatmaps of PathoGraph representations (adjacency matrices of PathoGraphs) for the ADNI and PPMI test sets, respectively. Notably, because brain graphs in ABIDE and ADHD-200 lack ROI labels, it is not feasible to obtain the correlation patterns of specific subgraphs from their PathoGraph visualizations. Therefore, we visualize the PathoGraph representations for the ADNI and PPMI datasets. In both ADNI and PPMI datasets, the representations show significant difference across diagnostic groups (AD, MCI, and NC in ADNI; PD, Prodromal, and NC in PPMI). These results demonstrate BrainPoG can effectively reveal disease-specific alterations.

\subsection{Additional Results of Ablation Study}\label{appendix_ab_results}

Tables~\ref{Tab:ab1_result2} and \ref{Tab:ab1_result3} respectively give the model performance (AUC and F1) and efficiency (running time and memory) of BrainPoG and its variants on four datasets. Table~\ref{Tab:ab2_results} shows the efficiency of models using different dimensionality reduction methods.

\begin{table*}[!htbp]
	\centering
	\caption{ Model performance (AUC and F1) of
BrainPoG and its variants on four datasets (\%).}
	\label{Tab:ab1_result2}
	\renewcommand\arraystretch{1}
    \setlength{\tabcolsep}{3pt} % Adjust column separation	
    \small
    \resizebox{\textwidth}{!}{
    \begin{tabular}{ccc|cc|cc|cc}
    \toprule
    \multirow{2}{*}{Method} & \multicolumn{2}{c|}{ADNI} &\multicolumn{2}{c|}{PPMI} &\multicolumn{2}{c|}{ABIDE} &\multicolumn{2}{c}{ADHD-200} \\\cmidrule{2-9}
    & AUC&F1&AUC&F1&AUC&F1&AUC&F1\\
    \midrule
    w/o Pattern Filter& 75.26$\pm$14.78 &47.33$\pm$15.05 &92.09$\pm$2.33&92.08$\pm$2.68& 74.89$\pm$17.67&56.88$\pm$23.04&99.81$\pm$0.02&96.04$\pm$3.49
    \\
    w/o Feature Distillation &54.49$\pm$4.47 &29.66$\pm$2.67&60.77$\pm$11.55 &21.62$\pm$2.76 &66.04$\pm$6.04 &50.10$\pm$9.77&71.63$\pm$5.48&62.47$\pm$3.66
    \\
    w/o Pattern Filter \& Feature Distillation& 55.65$\pm$5.40&23.85$\pm$2.52&58.01$\pm$4.90&24.46$\pm$ 5.55&53.96$\pm$4.94&41.93$\pm$10.49&73.64$\pm$5.96&64.52$\pm$7.04
      \\ 
     \midrule
     \textbf{BrainPoG} &92.23$\pm$2.83&69.86$\pm$8.21
     &92.00$\pm$1.50&81.22$\pm$1.66
  &98.92$\pm$0.67&93.13$\pm$2.49
&97.77$\pm$2.19&93.03$\pm$3.27
     \\
    \bottomrule
    \end{tabular}}
\end{table*}

\begin{table*}[!htbp]
	\centering
	\caption{ Model efficiency (running time and memory usage) of
BrainPoG and its variants on four datasets.}
	\label{Tab:ab1_result3}
	\renewcommand\arraystretch{1}
    \setlength{\tabcolsep}{3pt} % Adjust column separation	
    \small
    \resizebox{\textwidth}{!}{
    \begin{tabular}{ccc|cc|cc|cc}
    \toprule
    \multirow{2}{*}{Method} & \multicolumn{2}{c|}{ADNI} &\multicolumn{2}{c|}{PPMI} &\multicolumn{2}{c|}{ABIDE} &\multicolumn{2}{c}{ADHD-200} \\\cmidrule{2-9}
    & Runing time&Memory& Runing time&Memory&Runing time&Memory&Runing time&Memory\\
    \midrule
    w/o Pattern Filter& 0.0047s/epoch&48MB &0.0033s/epoch&26MB& 0.0041s/epoch&269MB& 0.0031s/epoch&121MB
    \\
    w/o Feature Distillation & 0.0051s/epoch &59MB&0.0029s/epoch &21MB &0.0044s/epoch&237MB&0.0029s/epoch&28MB
    \\
    w/o Pattern Filter \& Feature Distillation& 0.0055s/epoch&110MB&0.0029s/epoch&27MB&0.0051s/epoch&325MB&0.0031s/epoch&124MB
      \\ 
     \midrule
     \textbf{BrainPoG} & 0.0046s/epoch&23MB
  &0.0029s/epoch&17MB
    &0.0030s/epoch&99MB
    &0.0030s/epoch&23MB
     \\
    \bottomrule
    \end{tabular}}
\end{table*}

\begin{table*}[!htbp]
	\centering
	\caption{Efficiency of models using different dimensionality reduction methods.}
	\label{Tab:ab2_results}
	\renewcommand\arraystretch{1}
    \setlength{\tabcolsep}{3pt} % Adjust column separation	
    \small
    \resizebox{\textwidth}{!}{
    \begin{tabular}{cccc|ccc|ccc|ccc}
    \toprule
    \multirow{2}{*}{Method} & \multicolumn{3}{c|}{ADNI} &\multicolumn{3}{c|}{PPMI} &\multicolumn{3}{c|}{ABIDE} &\multicolumn{3}{c}{ADHD-200} \\\cmidrule{2-13}
    & Parameter No. & Running time&Memory&Parameter No. & Running time&Memory& Parameter No. & Running time&Memory& Parameter No. & Running time&Memory\\
    \midrule
    PCA& 34K&0.0030s/epoch&17MB
    &17K&0.0031s/epoch&19MB
    &68K&0.0027s/epoch&25MB
    &33K&0.0030s/epoch&17MB
    \\
    SVD& 39K&0.0047s/epoch&19MB
    &17K&0.0030s/epoch&19MB
    &67K&0.0029s/epoch&25MB
    &33K&0.0030s/epoch&17MB
    \\
     RFS& 40K&0.0052s/epoch&53MB
    &17K&0.0029s/epoch&20MB
    &67K&0.0046s/epoch&26MB
    &33K&0.0029s/epoch&35MB\\
    
    Autoencoder& 40K&0.0061s/epoch&127MB
    &17K&0.0031s/epoch&22MB
    &67K&0.0071s/epoch&526MB
    &33K&0.0029s/epoch&98MB
       
      \\ 
     \midrule
     \textbf{BrainPoG} & 227K& 0.0046s/epoch&23MB
     &11K&0.0029s/epoch&17MB
     &415K&0.0030s/epoch&99MB
     &140K&0.0030s/epoch&23MB
     \\
    \bottomrule
    \end{tabular}}
\end{table*}

\newpage
\paragraph{Effectiveness of Two Components in Pathological Feature Distillation Module.}\label{abltion_pathofeature}

To validate the effectiveness of noise-feature dropping and pathological feature augmentation in our pathological feature distillation module, we perform an ablation study in which each component is removed in turn. The results are reported in Table~\ref{Tab:ab1_result_noise_augmentation}. `\textit{w/o Noise Dropping}' denotes the model without the noise-feature dropping component, and `\textit{w/o Feature Augmentation}' indicates the model without the pathological feature augmentation component. Across all four datasets, removing pathological feature augmentation yields a marked performance drop, indicating that this strategy effectively enhances group-discriminative representations by amplifying group-specific features. Omitting noise-feature dropping likewise degrades accuracy and worsens efficiency, underscoring its role in reducing redundancy and preventing the model from learning disease-irrelevant features.

\begin{table*}[!htbp]
	\centering
	\caption{ Model performance (ACC (\%)) and efficiency (parameter number) of BrainPoG and its variants on four datasets.}
	\label{Tab:ab1_result_noise_augmentation}
	\renewcommand\arraystretch{1}
    \setlength{\tabcolsep}{3pt} % Adjust column separation	
    \small
    \resizebox{\textwidth}{!}{
    \begin{tabular}{ccc|cc|cc|cc}
    \toprule
    \multirow{2}{*}{Method} & \multicolumn{2}{c|}{ADNI} &\multicolumn{2}{c|}{PPMI} &\multicolumn{2}{c|}{ABIDE} &\multicolumn{2}{c}{ADHD-200} \\\cmidrule{2-9}
    & ACC&Parameter No.& ACC&Parameter No.&ACC&Parameter No.&ACC&Parameter No.\\
    \midrule
    w/o Noise Dropping& 78.65$\pm$4.44&257K &87.34$\pm$2.05&65K & 73.35$\pm$14.72 &1.7M& 88.24$\pm$2.76&164K
    \\
    w/o Feature Augmentation & 45.72$\pm$5.71 &245K&45.50$\pm$4.91 &63K &54.51$\pm$2.77 &1.6M&59.68$\pm$5.39&145K
      \\ 
     \midrule
     \textbf{BrainPoG} & 83.31$\pm$4.90 & 227K&82.90$\pm$1.62&11K & 93.16$\pm$2.46 &415K&93.03$\pm$3.27&140K
     \\
    \bottomrule
    \end{tabular}}
\end{table*}

\subsection{Additional Results of Hyperparameter Study}

\begin{figure*}[ht]
    \centering
    \begin{subfigure}{0.4\textwidth}
        \includegraphics[width=\linewidth]{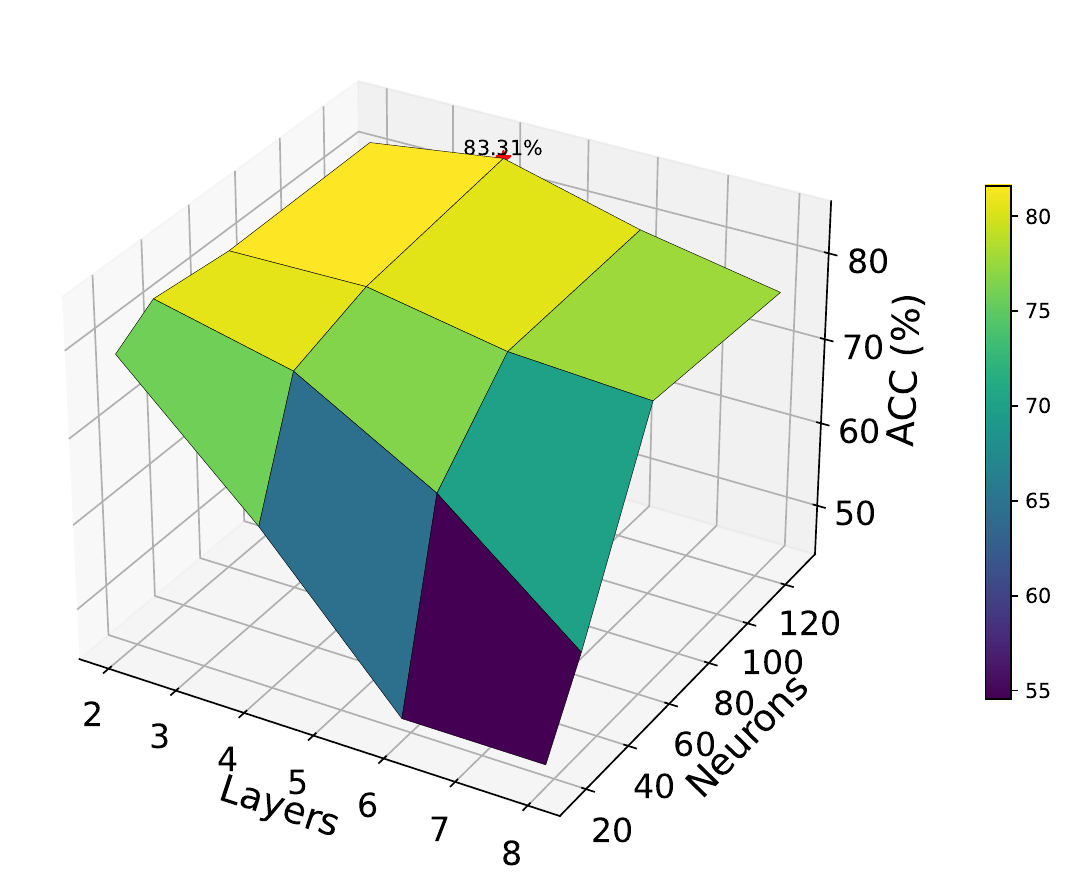}
        \caption{ADNI}
        \label{fig:sub1_hypara_layer}
    \end{subfigure}
   % \hfill
    \begin{subfigure}{0.4\textwidth}
        \includegraphics[width=\linewidth]{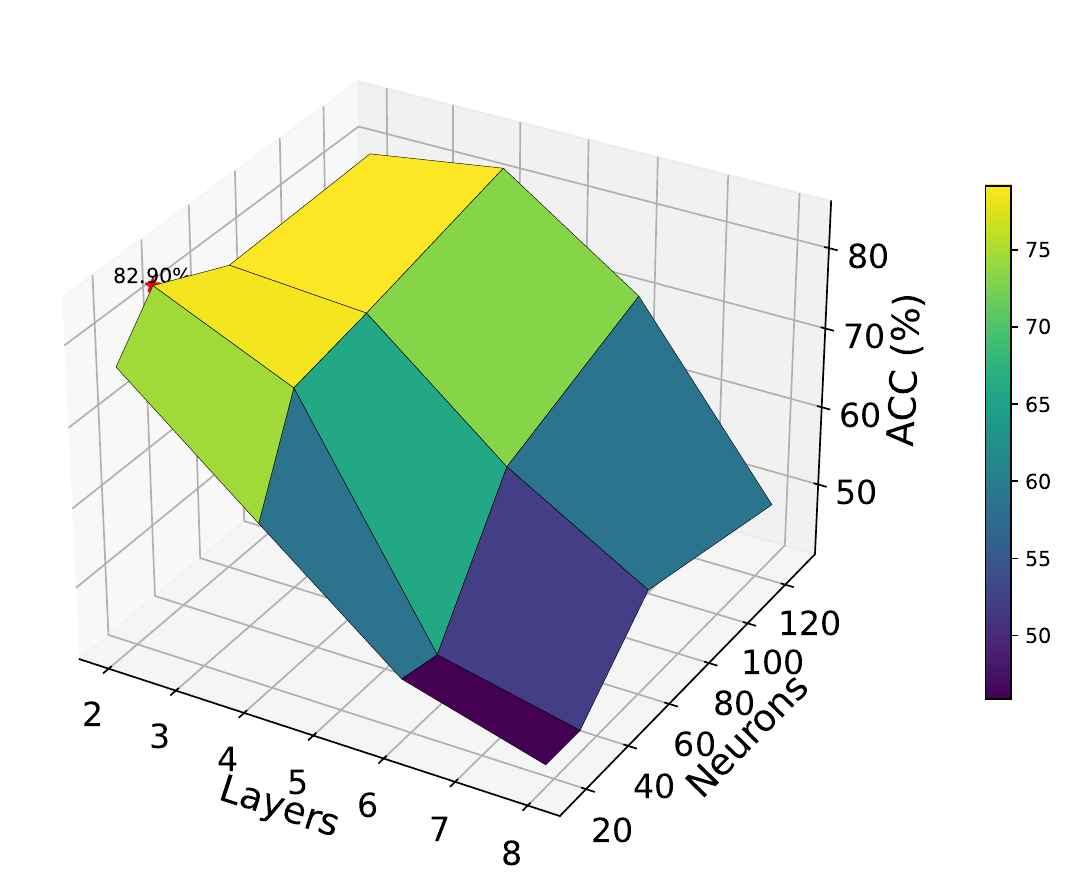}
        \caption{PPMI}
        \label{fig:sub2_hypara_layer}
    \end{subfigure}
   % \hfill
    \begin{subfigure}{0.4\textwidth}
        \includegraphics[width=\linewidth]{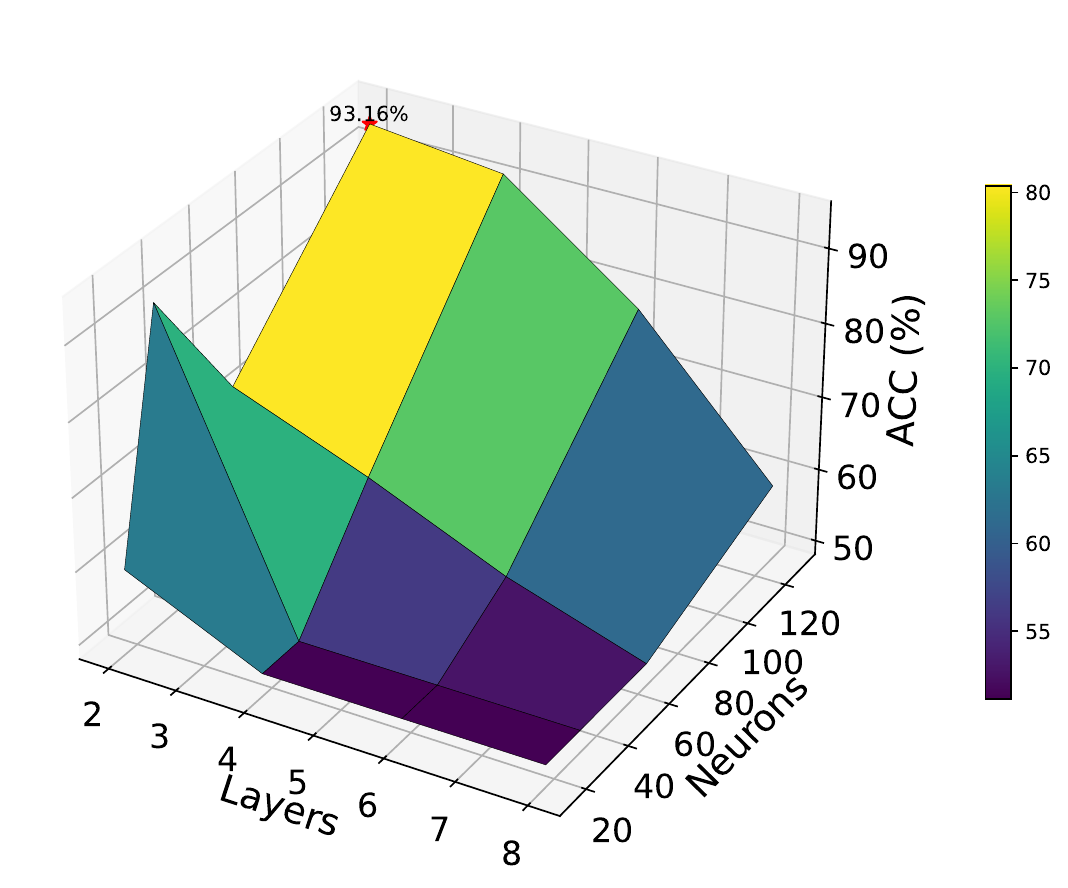}
        \caption{ABIDE}
        \label{fig:sub3_hypara_layer}
    \end{subfigure}
    \begin{subfigure}{0.4\textwidth}
        \includegraphics[width=\linewidth]{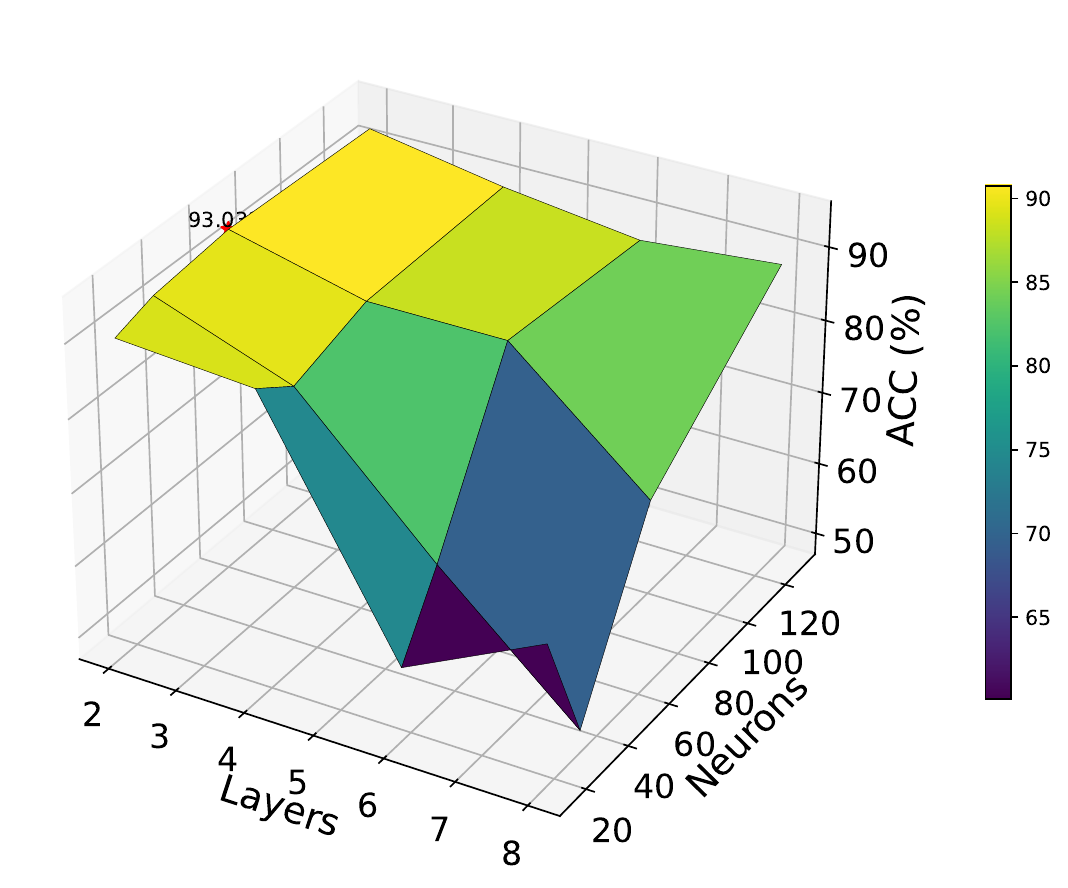}
        \caption{ADHD-200}
        \label{fig:sub4_hypara_layer}
    \end{subfigure}
    \caption{Results (ACC) under different numbers of layers and neurons.}
    \vspace{-1\baselineskip} 
    \label{fig:hyper_appendix_layer_nueron}
\end{figure*}

\paragraph{Experimental Results under Different Numbers of Layers and Neurons.}
To investigate model performance under different configurations and identify the optimal setting, we conduct experiments with varying numbers of GCN layers and neurons. The results are presented in Figure~\ref{fig:hyper_appendix_layer_nueron}.

\paragraph{Model Performance under Different Community Numbers.}

\begin{figure*}[ht]
    \centering
    \begin{subfigure}{0.4\textwidth}
        \includegraphics[width=\linewidth]{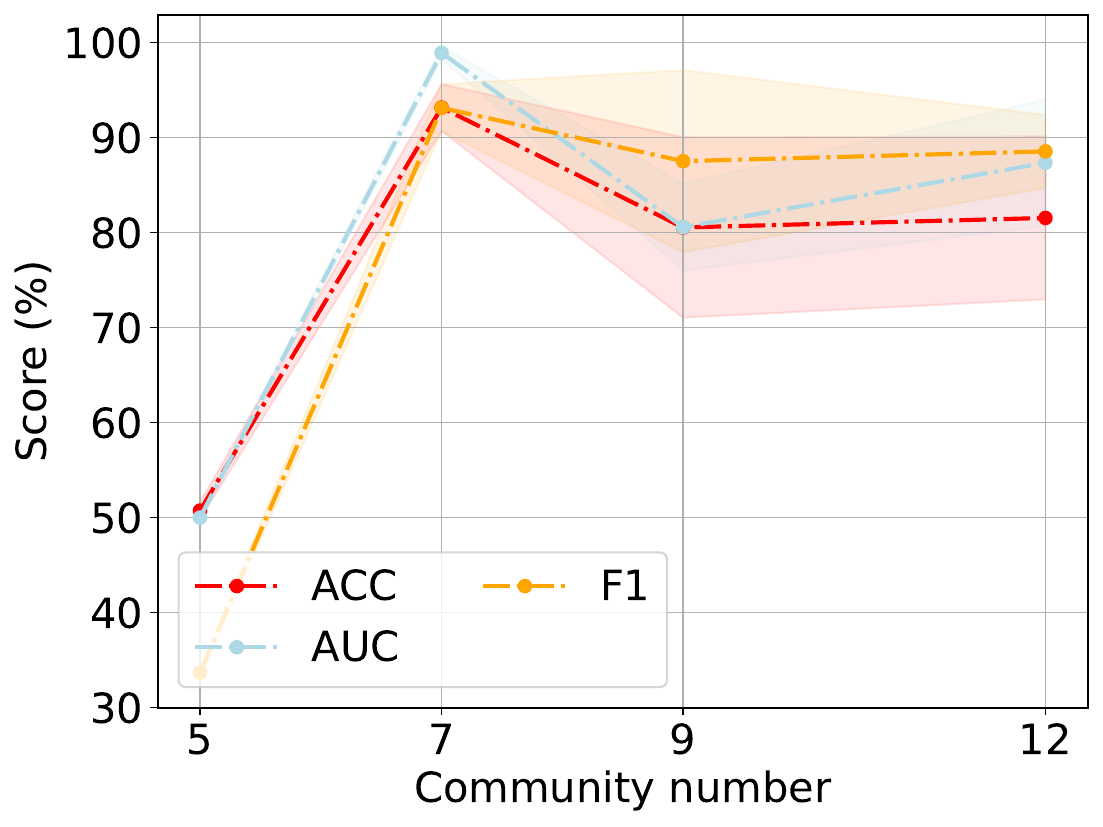}
        \caption{ABIDE}
        \label{fig:sub3_hypara_community}
    \end{subfigure}
    \begin{subfigure}{0.4\textwidth}
        \includegraphics[width=\linewidth]{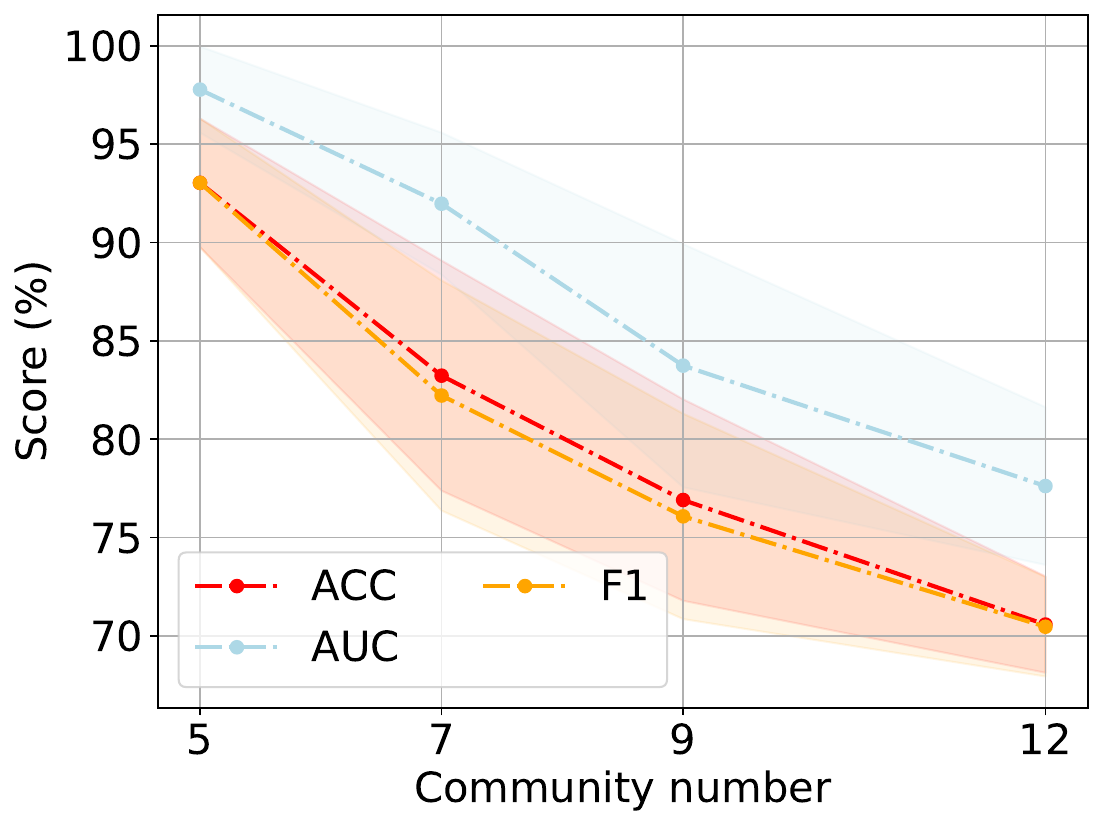}
        \caption{ADHD-200}
        \label{fig:sub4_hypara_community}
    \end{subfigure}
    \caption{Model performance under different community numbers in ABIDE and ADHD-200 datasets.}
    \vspace{-1\baselineskip} 
    \label{fig:hyper_appendix_community}
\end{figure*}

As mentioned before, we use spectral clustering to obtain subgraphs for brain graphs in ABIDE and ADHD-200 datasets. As spectral clustering is unsupervised method, different community numbers can be defined. To investigate the influence of community number on model performance, we conduct experiments with different community numbers. Figure~\ref{fig:hyper_appendix_community} shows the model performance under different community numbers. As shown, the community number has a substantial impact on the model performance. This indicates the importance of subgraph division, reflecting functional modules.

\end{document}